\documentclass{vldb}
\usepackage{graphicx}
\usepackage{balance}
\usepackage[affil-it]{authblk}

\usepackage{array}

\usepackage[normalem]{ulem}
\usepackage[hyphens]{url}

\begin{document}

\title{CuMF\char`_SGD: Fast and Scalable Matrix Factorization}
\renewcommand\Authfont{\fontsize{12}{14.4}\selectfont}
\renewcommand\Affilfont{\fontsize{10}{10.8}\itshape}


%
%
\font\authorfont=cmr12 at 12pt

\author[1]{Xiaolong Xie\thanks{Work done when the author was with IBM.}\hspace{0.15cm}}
\author[2]{\authorfont Wei Tan}
\author[2]{\authorfont Liana L. Fong}
\author[1]{\authorfont Yun Liang}
\vspace{-0.2cm}
\affil[1]{Center for Energy-efficient Computing and Applications, School of EECS, Peking University, China \hspace{1cm}
{xiexl\_pku, ericlyun@pku.edu.cn}}
\affil[2]{IBM Thomas J. Watson Research Center, New York, U.S.A

wtan, llfong@us.ibm.com}
\vspace{-0.2cm}

\maketitle
%

\begin{abstract}
Matrix factorization (MF) has been widely used in recommender systems, database systems, topic modeling, word embedding and others. Stochastic gradient descent (SGD) is popular in solving MF problems because it can deal with large data sets and is easy to do incremental learning. 
We observed that SGD for MF is memory bound. Meanwhile, single-node CPU systems with caches perform well only for small data sets; distributed systems have higher aggregated memory bandwidth but suffer from relatively slow network connection. This observation inspires us to accelerate MF by utilizing GPUs's high memory bandwidth and fast intra-node connection.

We present \textbf{cuMF\_SGD}, a CUDA-based SGD solution for large-scale MF problems. On a single GPU, we design two workload scheduling schemes (batch-Hogwild! and wavefront-update) that fully exploit the massive amount of cores. Especially, batch-Hogwild!, a vectorized version of Hogwild!, overcomes the issue of memory discontinuity. We develop highly-optimized kernels for SGD update, leveraging cache, warp-shuffle instructions, half-precision floats, etc. We also design a partition scheme to utilize multiple GPUs while addressing the well-known convergence issue when parallelizing SGD. Evaluations on three data sets with only one Maxwell or Pascal GPU show that cuMF\_SGD runs \textbf{3.1X-28.2X} as fast compared with state-of-art CPU solutions on 1-64 CPU nodes. Evaluations also show that cuMF\_SGD scales well with multiple GPUs on large data sets. Finally, we believe that the lessons learned from building cuMF\_SGD are applicable to other machine learning algorithms on, e.g., (1) embedding layers in deep learning and (2) bipartite graph.

\end{abstract}

\section{Introduction}\label{sec:introduction}

\begin{figure}[h]
\centering
\vspace{-0.1cm}
\includegraphics[scale=0.49,angle=0,natwidth=610]{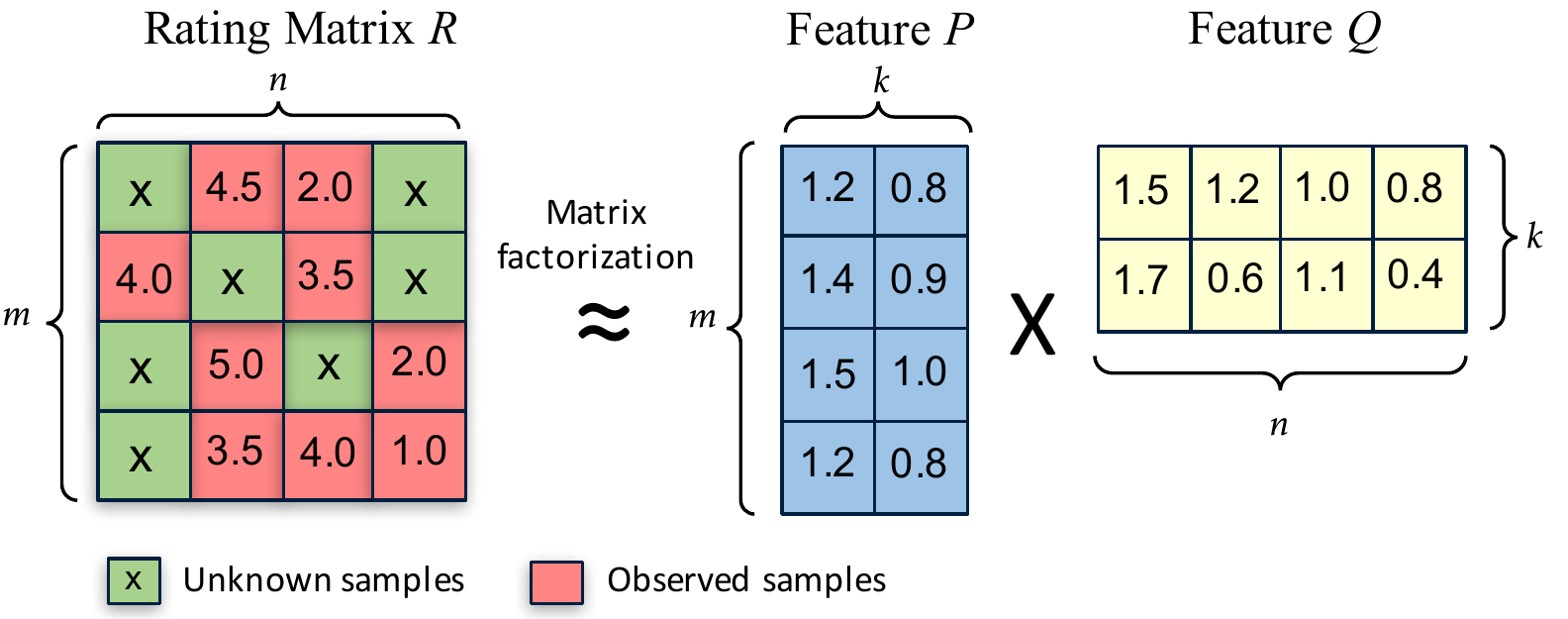}
\vspace{-0.25cm}
\caption{An example of matrix factorization where $m$=4, $n$=4, $k$=2.}\label{figure:basic}
\end{figure}

Matrix factorization (MF) has been widely used in recommender systems~\cite{computer09mf} by many companies (i.e. Amazon, Netflix, Facebook~\cite{facebook} and Spotify). It can also be used in topic modeling, word embedding~\cite{glove2014}, database system~\cite{sarwat2014database}, and has a natural connection to the embedding layers in deep neural network. Let us use the recommender system as example. Figure~\ref{figure:basic} shows a rating matrix $R$ of $m \times n$, with sparse ratings from $m$ users to $n$ items. We assume that $R$ can be factorized into the multiplication of two low-rank feature matrices $P$ ($m \times k$) and $Q$ ($k \times n$), such that $R\approx P \times Q$. The derived feature matrices $P$ and $Q$ can be used to predict the missing ratings in $R$, or as features of corresponding users/items in downstream machine learning tasks. Matrix factorization often involves large data sets. For example, the number of users/items may range from thousands to hundreds-of-millions, and the number of observed samples in $R$ may range from millions to tens-of-billions~\cite{facebook}. Therefore, there is a need to scale and speed up large-scale matrix factorization.
\vspace{-0.2cm}

There are mainly three algorithms to solve matrix factorization, i.e., coordinate gradient descent(CGD), alternate least square (ALS), and stochastic gradient descent (SGD). As previous works show that CGD is prone to reach local optima~\cite{tist15libmf}, we do not focus on it in this paper. ALS is easy to parallelize, and able to deal with implicit feedback such as purchase history~\cite{computer09mf}. Meanwhile, SGD usually converges faster because it is not as computation-intensive as ALS; SGD is also more applicable in incremental learning settings where new ratings are continuously fed in. With our previous work already tackled ALS~\cite{hpdc16wei}, we focus on SGD in this paper.

The optimization work on matrix factorization contains two streams: algorithm and system. The algorithmic stream tries to optimize update schemes such as learning rate in gradient descent, in order to \textbf{reduce the number of epochs} (an epoch is a full pass through the training set) needed to converge~\cite{libmf++}. The system stream tries to accelerate the computation, in order to \textbf{run each epoch faster} ~\cite{hpdc16wei,tist15libmf,dsgd12,yu2012scalable,vldb14nomad,recht2011hogwild}. We focus the system stream and the proposed techniques can be combined with other algorithmic optimizations. Our research is based on the following observations. 

\textbf{Observation 1. MF with SGD is memory bound.}

When solving MF with SGD, in each step of an epoch, we randomly select one observed sample, read the corresponding features $p$ and $q$, do an inner product of them, update $p$ and $q$, and eventually write them back (details to be given in Section~\ref{sec:sgd}). Obviously, the \textbf{compute operations per byte is very low}. For example,, if we use single-precision (4 byte float) and $k=128$, one SGD update involves 2432 bytes memory access (384 bytes for $R$ + 2048 bytes for $p \& q$ read/write) and 1536 floating point operations (256 ops for dot product and 1280 ops for feature update). That is, the flops/byte ratio is $1536/2432\approx 0.63$.
To put this number into perspective, a modern processor can achieve 600 Gflops/s and memory bandwidth 60 GB/s. This means that the operation's flops/byte ratio has to be as large as $600/60=10$ to saturate the computation units. The low flops/byte ratio of MF with SGD indicates that, its performance is bounded by memory bandwidth.

State-of-art SGD-based MF solutions are based on either shared-memory multi-threading~\cite{tist15libmf} or distributed systems~\cite{vldb14nomad}. We observed that neither of them is capable of offering sustained high memory bandwidth to MF.

\textbf{Observation 2.  Single-node CPU systems with caching achieve high memory bandwidth only for small data sets. Distributed systems have higher theoretical bandwidth, but handicapped by the relatively weak network connection.}

\begin{figure}[h]
\centering
\includegraphics[scale=0.3,angle=0]{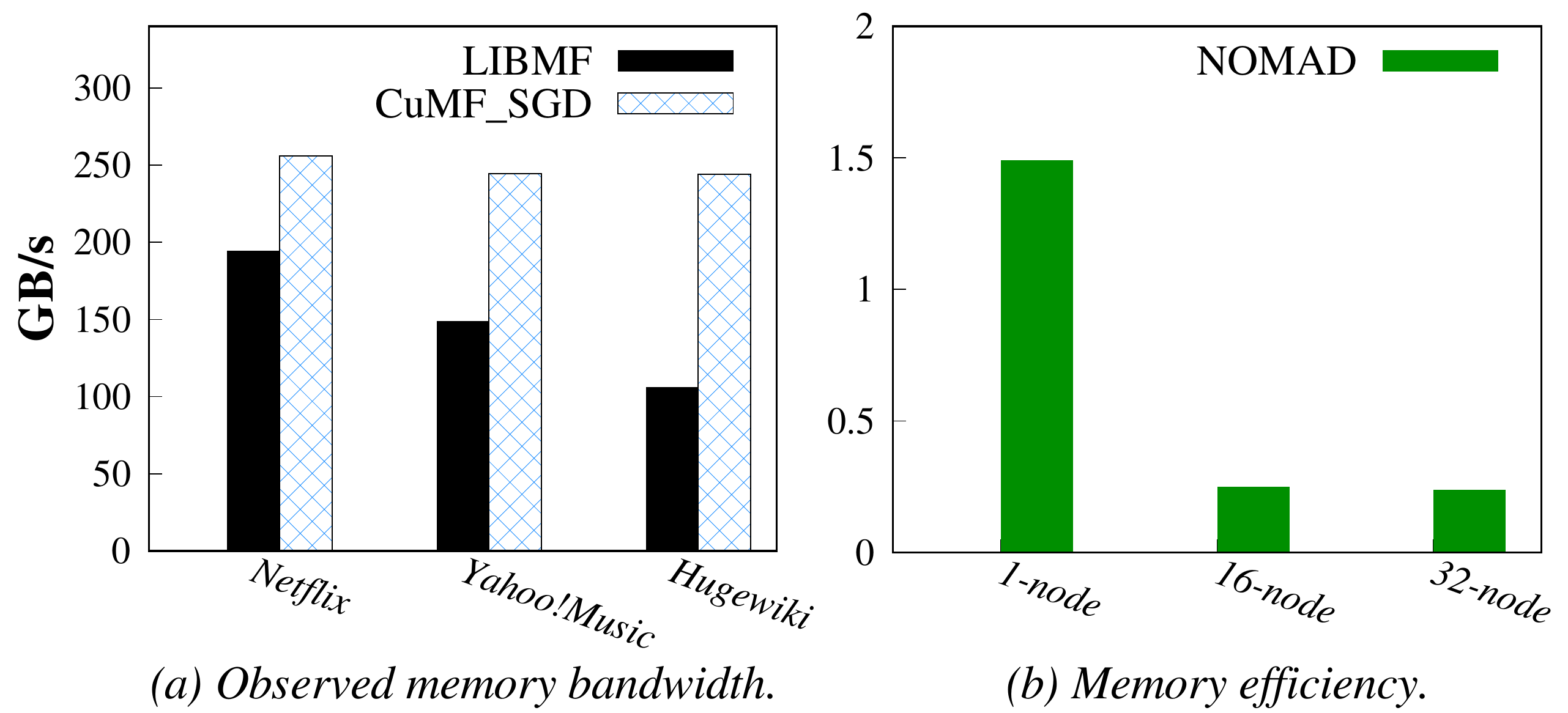}
\caption{(1) The observed memory bandwidth of LIBMF drops when solving large data sets. (b) NOMAD achieves lower memory efficiency when scaling to multiple nodes.}\label{figure:mot}
\vspace{-0.4cm}
\end{figure}

Shared-memory CPU systems~\cite{tist15libmf,recsys15fast,sigmod11hash} rely heavily on cache to achieve high memory throughput. To understand this, we evaluate a single-node and shared-memory MF library \textit{LIBMF}~\cite{tist15libmf} with three data sets (details shown in Section~\ref{sec:exp}). As seen from Figure~\ref{figure:mot}(a), on the small \textit{Netflix} data set, LIBMF achieves an observed 194 GB/s bandwidth\footnote{Observed bandwidth means the aggregated bandwidth offered by DRAM and cache that is observed by the compute unit.}, much larger than the actual system DRAM bandwidth ($\sim60$ GB/s). However, on a much larger \textit{Hugewiki} data set, the achieved memory bandwidth drops by 45\%, to 106 GB/s. This is because that Hugewiki data set is large enough to vanish a lot of data locality. This evaluation demonstrates that single-node CPU systems cannot achieve high memory bandwidth when solving large-scale MF problems.

Distributed systems are frequently used to accelerate time-consuming applications~\cite{yang2016husky,low2012distributed}. Distributed systems can aggregate the memory bandwidth and caches on multiple nodes. However, SGD is inherently sequential and consequently different nodes need to reconcile the parameters at the end of each epoch. As a result, despite of the high aggregated memory bandwidth, the performance is bounded by the limited network bandwidth between computer nodes. Figure~\ref{figure:mot}(b) evaluates \textit{NOMAD}~\cite{vldb14nomad}, a distributed MF system. We measure its memory efficiency which is defined as the ratio of achieved memory bandwidth to the aggregated memory bandwidth of all nodes. The memory efficiency of NOMAD rapidly decreases when scale to multiple nodes, due to the slow over-network communication.

\textbf{Observation 3. GPUs are with much higher memory bandwidth and enjoy fast inter-device connection within a single node, making GPU an ideal candidate to accelerate MF.}

GPUs are widely used to accelerate applications with large data sets, e.g., machine learning and data base systems~\cite{vldb15skyline,zhang2015mega,wang2014concurrent}. Observations 1-2 inspire us to resort to GPUs for the following reasons. Firstly, GPUs are with high off-chip memory bandwidth. For example, NVIDIA Maxwell GPUs has theoretical 360 GB/s off-chip memory bandwidth~\cite{maxwell} and the newer generation Pascal GPUs provide 780 GB/s~\cite{pascal}. These are several times to an-order-of-magnitude higher than CPUs. Secondly, GPUs do not rely on cache to reduce \textbf{latency}; instead, they rely on thousands of concurrent threads running on hundreds of cores to achieve high \textbf{throughput}~\cite{vldb16gpu, zhou2016generic}. Therefore, unlike CPUs, GPUs' achieved memory bandwidth does not degrade when working data set exceeds cache capacity, as shown in Figure~\ref{figure:mot}(a). Thirdly, multiple GPUs can enjoy very fast interconnect in a node. For example, the recent NVLink~\cite{nvlink} can offer 160 GB/s per-GPU bandwidth to CPU and peer-GPUs. This is much faster than PCIe 3.0 (32GB/s for 16x) and InfiniBand (6.8 GB/s for 56Gb/s FDR). 

\textbf{Proposed solution: cuMF\_SGD to accelerate MF by utilizing one or multiple GPUs' high memory bandwidth and intra-node connection.}

Parallelizing SGD on GPUs is challenging. Due to the architecture distinct, simply mapping CPUs' algorithms to GPUs will lead to extremely low performance and suboptimal resources usage~\cite{pldi2010gpu,zhao2013parallelizing}. Moreoever, SGD is inherently serial, studies~\cite{ipdpsw15japan} have shown that existing MF solutions do not scale well using merely 30 threads. Hence, to accelerate SGD to MF on GPUs, comprehensive understanding of GPU architectural features and novel algorithms are needed.

We present \textbf{cuMF\_SGD}\footnote{\url{http://github.com/cumf/cumf_sgd/}}, a fast and scalable SGD-based MF solution on GPUs. Inspired by the lock-free~\cite{recht2011hogwild} and the block-based~\cite{sigkdd11gemulla, tist15libmf} approaches, and given the separated CPU/GPU memory space, cuMF\_SGD adopts a hybrid two-level execution scheme. (1) At the top level, the rating and feature matrices are partitioned into blocks and distributed to multiple GPUs. (2) At the second level, each GPU does SGD with its own partition. One GPU only synchronizes with other GPUs through the shared CPU memory, when it completes the processing of its partition. Within each GPU, the GPU-local data partition is further distributed to the hundreds of thread blocks. Each thread block processes SGD updates with highly-optimized vector operations. By this means, cuMF\_SGD is able to scale to massive threads on multiple GPUs and perform highly-vectorized operations.

The contributions of this paper are as follows:
\begin{itemize}
\item \textit{Workload characterization}. We identify that SGD-based MF is bounded by memory bandwidth and synchronization overhead, instead of computation. We also identify the two challenges in implementing MF on GPUs using SGD, i.e., workload partitioning to avoid conflict and efficient update to exploit GPU hardware.

\item \textit{Optimization on a single GPU}. We design two ways to partition and schedule the work within a single GPU, i.e., (1) matrix blocking-based algorithm with lightweight, wave-based scheduling policy, and (2) batch-Hogwild! which can be seen as a mini-batch version of the original Hogwild! algorithm. Besides the scheduling schemes, we also develop highly optimized GPU kernels for SGD update. We leverage the architectural features such as cache, warp shuffle instructions, and half-precision floats with 16 bits.

\item \textit{Deal with big data on multiple GPUs}. We design a scheme to partition large data sets and solve them on multiple GPUs. We overlap data transfer and computation to minimize the execution time. We also analyze the relation between the number of partitions and the number of workers, which impacts the randomness of the SGD update, and ultimately the convergence speed.

\item \textit{Evaluation}. We implement cuMF\_SGD in a shared-memory system with multiple GPUs. Experimental results show that, cuMF\_SGD with one GPU is \textbf{3.1X-28.2X} as fast compared with state-of-art CPU solutions on 1-64 CPU nodes. CuMF\_SGD is also able to scale to multiple GPUs and different generations GPUs.

\end{itemize}

The remaining of this paper is organized as follows. Section~\ref{sec:background} analyzes the targeted GPU architecture, the workload of SGD for MF, and its parallelization schemes. Section~\ref{sec:single} presents the single-GPU cuMF\_SGD, including the two scheduling schemes and the GPU kernel. Section~\ref{sec:scale} discusses how to solve large-scale problems by matrix partitioning. Section~\ref{sec:exp} presents and discusses experiment results. Section~\ref{sec:related} discusses the related work and Section~\ref{sec:conclusion} concludes this paper.

\section{Background}\label{sec:background}
This section first briefly introduces the GPU architecture, and the SGD algorithm for matrix factorization. Then, we discuss the parallelism schemes for MF with SGD. We argue that, the current lock-free and matrix blocking schemes are not scalable on massive GPU cores. This leads to the innovations to be presented in Sections 3 and 4.
\subsection{The Compute Architecture}
\begin{figure}[h]
\centering
\includegraphics[scale=0.68,angle=0]{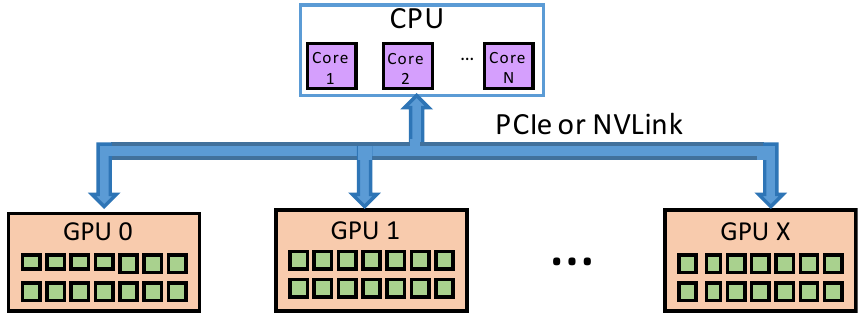}
\caption{The compute architecture for cuMF\_SGD.}\label{figure:arch}
\end{figure}

To overcome the limited memory bandwidth of a single CPU node and limited network bandwidth of distributed systems, we use a heterogeneous platform with GPUs, as shown in Figure~\ref{figure:arch}. GPUs have high intra-device memory bandwidth and are connected via PCIe or NVLink~\cite{nvlink} with high inter-device bandwidth. The CPUs take care of data pre-processing, data movement, and top-level workload scheduling, while the GPUs deal with feature update with massive parallelism.

GPUs are throughput-oriented processors~\cite{guide} with thousands of cores and high bandwidth memory (200-800 GB/s). To fully exploit the performance potential, GPU applications need to be carefully designed to exploit the data and computation parallelism. In the next two sub-sections we show that is non-trivial for SGD as SGD is inherently serial.

\subsection{Stochastic Gradient Descent}\label{sec:sgd}
Given $m$ users and $n$ items and a sparse rating matrix $R$, where $r_{u,v}$ indicates the preference or rating of $u_{th}$ user on $v_{th}$ item. The goal of matrix factorization is to train a $m \times k$ feature matrix $P$ and a $k \times n$ feature matrix $Q$ such that:
$$ \textbf{R} \approx \textbf{P} \times \textbf{Q} $$
The training process of matrix factorization is to minimize the following cost function:
$${\sum\limits_{u,v \in R} (r_{u,v} - \textbf{p}_u\textbf{q}_v)^2 + \lambda_p\mid\mid \textbf{p}_u\mid\mid^2 + \lambda_q\mid\mid \textbf{q}_v\mid\mid^2 } $$
where $\lambda_p$ and $\lambda_q$ are regularization parameters to avoid overfitting and $N$ is the number of non-zero samples in matrix $R$. 
The key idea of SGD is in every single step, randomly select one sample, e.g., $r_{u,v}$ from $R$, to calculate the gradient w.r.t to the following cost function:
$${ (r_{u,v} - \textbf{p}_u\textbf{q}_v)^2 + \lambda_p\mid\mid \textbf{p}_u\mid\mid^2 + \lambda_q\mid\mid \textbf{q}_v\mid\mid^2 } $$
Then update feature vectors with learning rate $\alpha$:
$$err_{u,v} = r_{u,v} - \textbf{p}_u\textbf{q}_v$$
$$\textbf{p}_u \gets \textbf{p}_u + \alpha(err_{u,v}\textbf{q}_v^T - \lambda_p\textbf{p}_u) $$
$$\textbf{q}_v \gets \textbf{q}_v + \alpha(err_{u,v}\textbf{p}_u^T - \lambda_q\textbf{q}_v) $$

\subsection{Parallelization Schemes}
\begin{figure}[h]
\centering
\includegraphics[scale=0.47,angle=0]{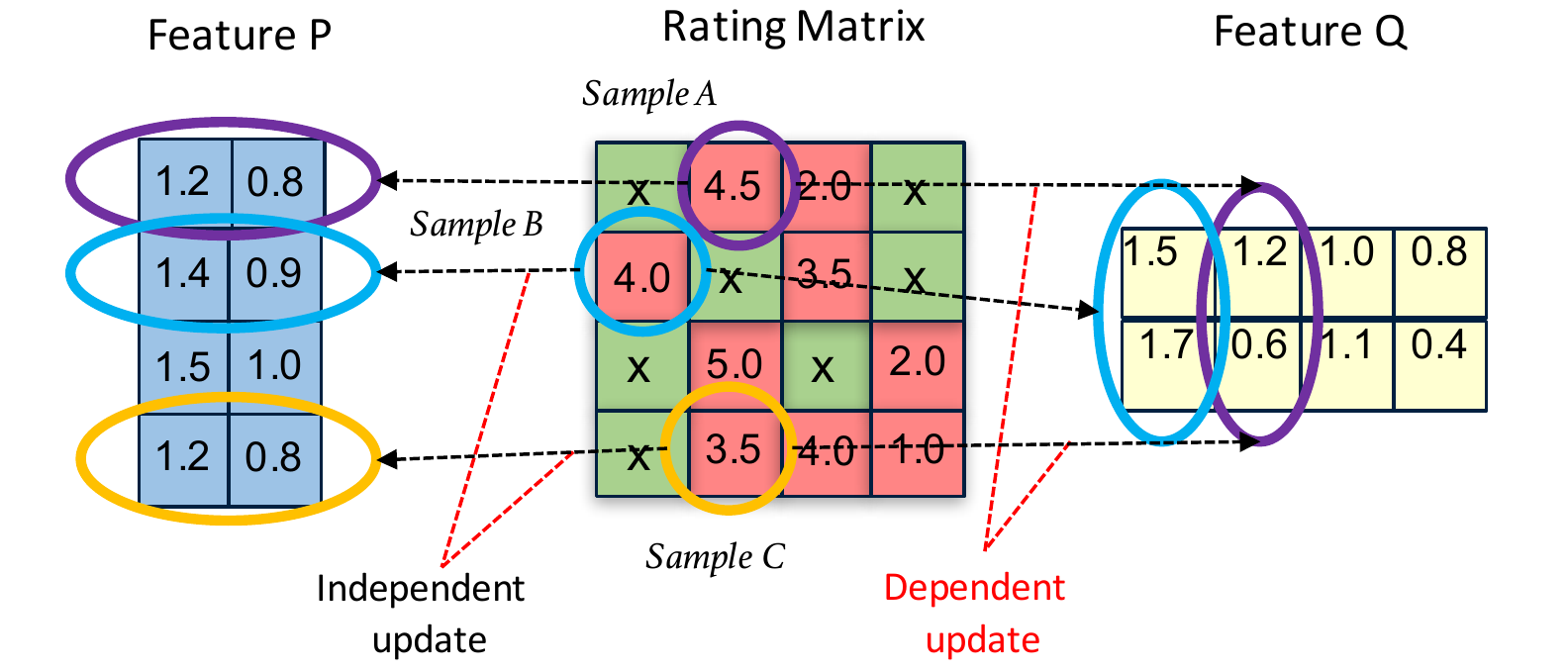}
\caption{Samples in different rows and columns are independent, e.g., samples \textit{A} and \textit{B}. Otherwise they are dependent, e.g. samples \textit{A} and \textit{C}.}
\label{figure:independent}
\end{figure}

SGD is inherently serial where each time one sample is selected to update. Given a data set with $N$ samples, an epoch (aka., iteration) involves executing $N$ updates one after other. Usually, it takes tens to hundreds of epochs to converge. To accelerate this process, it was observed that two samples can update in parallel if they are neither in same row nor same column. In this paper, we call these samples as "independent samples". Figure~\ref{figure:independent} shows an example. \textit{Sample A} (row 0, column 1) and \textit{Sample B} (row 1, column 0) are \textbf{independent} as they are associated with different rows in $P$ and different columns in $Q$. Meanwhile, \textit{Sample A} and \textit{Sample C} are \textbf{dependent} as they update the sample column 1 in $Q$. Ideally, SGD can be parallelized without losing accuracy, if we update independent samples in parallel, and update dependent samples sequentially~\cite{tist15libmf}.

To accelerate the SGD algorithm, how to partition the workloads to parallel workers becomes one major challenge. The efficiency of the workload scheduling scheme has a profound impact to the convergence speed. The workload scheduling policies in existing work ~\cite{tist15libmf,vldb14nomad,ipdpsw15japan,recht2011hogwild,sigkdd11gemulla} can be divided into two categories, \textbf{Hogwild!} and \textbf{matrix-blocking}.
\begin{figure}[h]
\centering
\includegraphics[scale=0.6,angle=0]{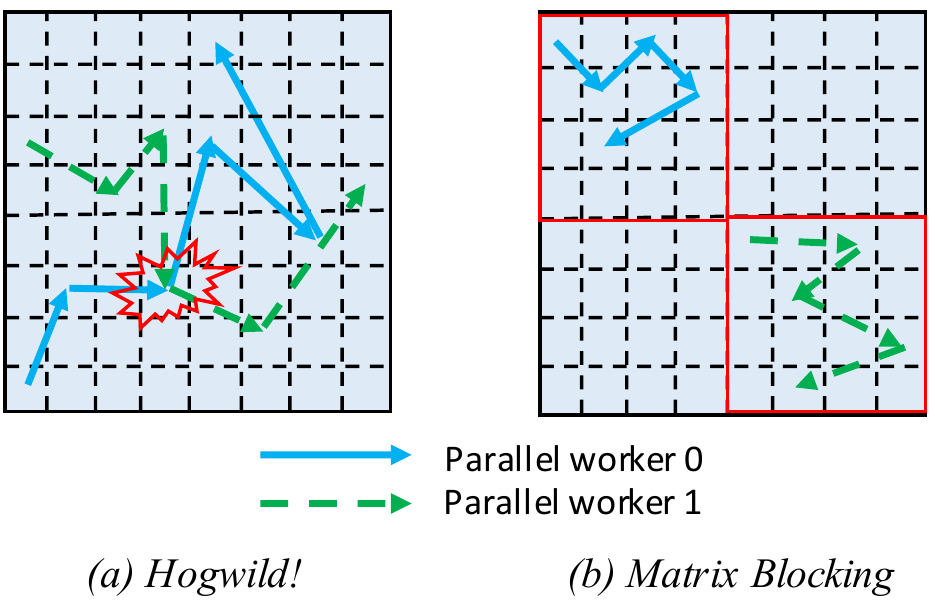}
\vspace{-0.2cm}
\caption{Two SGD parallelization schemes for MF: \textit{Hogwild!} and \textit{matrix-blocking}.}\label{figure:schema}
\vspace{-0.4cm}
\end{figure}

\textbf{Hogwild!}(Figure~\ref{figure:schema}(a)) is a lock-free approach to parallelize SGD\cite{recht2011hogwild}. In this paper, we call that a \textit{conflict} happens if two or more concurrent threads update samples in the same row or column at the same time. Intuitively, certain synchronization should be used to avoid conflicts. However, \textit{Hogwild!} observes that such synchronization is not needed when the $R$ is very sparse and the number of concurrent threads is much less than the number of samples. The intuition is that, when the aforementioned requirements are met, the probability of conflicts is very low and the incurred accuracy loss can be ignored. Based on the low synchronization overhead, it is used by some MF solutions. However, we need to enhance \textit{Hogwild!} in cuMF\_SGD in two aspects: (1) \textit{Hogwild!} assumes a global shared memory space, which is not feasible in our hybrid CPU/GPU setting where each GPU has its own memory space. We can only run \textit{Hogwild!} in a single GPU and need another layer of partition among multiple GPUs. (2) The vanilla \textit{Hogwild!} is not cache friendly because of random access. How to balance the randomness in SGD update and cache efficiency is an important issue at design time.

\textbf{Matrix-blocking} (Figure~\ref{figure:schema}(b)) divides the rating matrix into several sub-blocks, and sub-blocks that do not share rows or columns can update in parallel. Matrix-blocking is used by many recent work  \cite{vldb14nomad,tist15libmf,ipdpsw15japan,sigkdd11gemulla}. Matrix-blocking has advantage in that, it totally avoids conflict. However, in matrix-blocking parallel workers need to ask a global scheduler on which blocks to proceed next. This global scheduler has been shown not scalable to many-core architectures~\cite{ipdpsw15japan}. Hence, we need to enhance existing matrix-blocking schemes to scale to the many cores on GPUs.



\vspace{0cm}
\section{Single GPU Implementation}\label{sec:single}

This section presents how cuMF\_SGD solves MF with one GPU. We pre-assume that all required data resides in GPU device memory. We discuss multi-GPU implementation in Section~\ref{sec:scale}. We need to tackle two issues in single GPU. Section~\ref{sec:vec} discusses the issue of \textbf{computation optimization}, i.e., to optimize each individual SGD update by exploiting GPU hardware. Section~\ref{sec:scheduling} discusses \textbf{workload scheduling}, i.e., to distribute the many SGD updates to thousands of concurrent GPU threads.

\subsection{Computation Optimization}\label{sec:vec}
\begin{figure*}
\centering
\includegraphics[scale=0.45,angle=0]{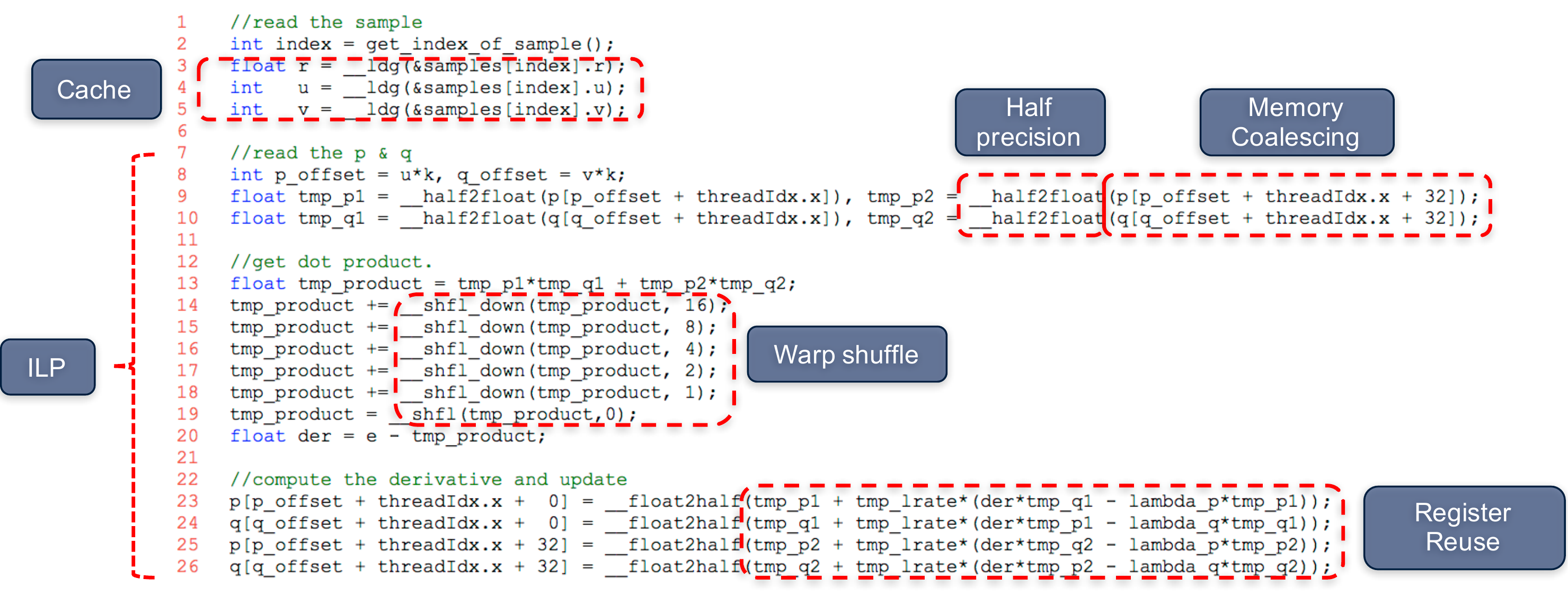}
\vspace{-0.25cm}
\caption{The exemplify code of computation part of cuMF\_SGD, where $k=64$. The used optimization techniques are highlighted.}\label{figure:code}
\vspace{-0.2cm}
\end{figure*}

In MF, one SGD update consists of four steps: 1) read one sample ($r_{u,v}$) from the rating matrix, 2) read two feature vectors  ($p_u$, $q_v$), 3) compute prediction error $r_{u,v} - p_uq_v$, and 4) update the features. Except the first step, other three steps are all vector operations at length $k$. $k$ is an input parameter and typically ranges from $\mathcal {O}(10)$ to $\mathcal {O}(100)$. On a CPU, a parallel worker can be a thread or process, where vector instructions such as SSE and AVX can be used to accelerate the computation. GPUs are SIMD architectures~\cite{icde16simd}, where a thread block is a vector group. Hence, in cuMF\_SGD, we use a thread block as a parallel worker. Figure~\ref{figure:code} shows a code snippet of the computational part of cuMF\_SGD, where we use $k=64$ as an example. We highlight the major optimization techniques in Figure~\ref{figure:code} and explain them in the following.

\textbf{Cache}. Since Fermi architectures, NVIDIA GPUs feature on-chip L1 cache and allows the programmers to control the cache behavior of each memory instruction (cache or bypass). While many GPU applications do not benefit from cache due to cache contention~\cite{iccad13xie}, some memory instructions may benefit from cache as the accessed data may be frequently reused in the future (temporal reuse) or by other threads (spatial reuse). Following the model provided by~\cite{iccad13xie}, we observe that the memory load of the rating matrix benefits from cache and use the intrinsic instruction \textit{\_\_ldg}~\cite{guide} to enable cache-assisted read.

\textbf{Memory coalescing}.
On GPUs, when threads within one warp access the data within one cache line, the access is coalesced to minimize the bandwidth consumption~\cite{pldi2010compile}. This is called memory coalescing. In cuMF\_SGD, the read/write of $P$ and $Q$ are carefully coalesced to ensure that consecutive threads access consecutive memory addresses.

\textbf{Warp shuffle}. Warp shuffle instructions~\cite{sc13shuffle} are used to compute the dot product $p \cdot q$  and broadcast the result. Compared with traditional shared-memory-based approaches, this warp shuffle-based approach performs better because: (1) warp shuffle instructions have extra hardware support, (2) register is faster than shared memory, and (3) no thread synchronization is involved. To exploit the warp shuffle feature, we fix the thread blocks size as warp size(32).

\textbf{ILP}. Modern GPUs support compiler-aided super scalar to exploit the instruction-level parallelism (ILP). In cuMF\_SGD, when $k>32$, a thread is responsible to process $k/32$ independent scalars. Hence, with awareness of the low-level architecture information, we reorder the instructions to maximize the benefit of ILP.

\textbf{Register usage}. Register file is an important resource on GPUs~\cite{cgo13sgemm}. As the total number of registers on GPUs are limited, while each thread uses too many registers, the register consumption may become the limitation to concurrency. In our case, we identify that the concurrency is only limited by the number of thread blocks of GPUs~\cite{guide}. Hence, we allocate as many as possible registers to each thread such that every reusable variable is kept in the fastest register file.

\vspace{-0.2cm}
\textbf{Half-precision}. As addressed before, SGD is memory bound. Most of the memory bandwidth is spent on the read/write to the feature matrices. Recently, GPU architectures support the storage of half-precision (2 bytes vs. 4 bytes of single-precision) and fast transformation between floating point and half-precision. In practice, after parameter scaling, half-precision is precise enough to store the feature matrices and do not incur accuracy loss. CuMF\_SGD uses half-precision to store feature matrices, which halves the memory bandwidth need when accessing feature matrices.

\subsection{Workload Scheduling}\label{sec:scheduling}
\subsubsection{Scalability issue of global scheduling}
\begin{figure}[h]
\centering
\includegraphics[scale=0.45,angle=0]{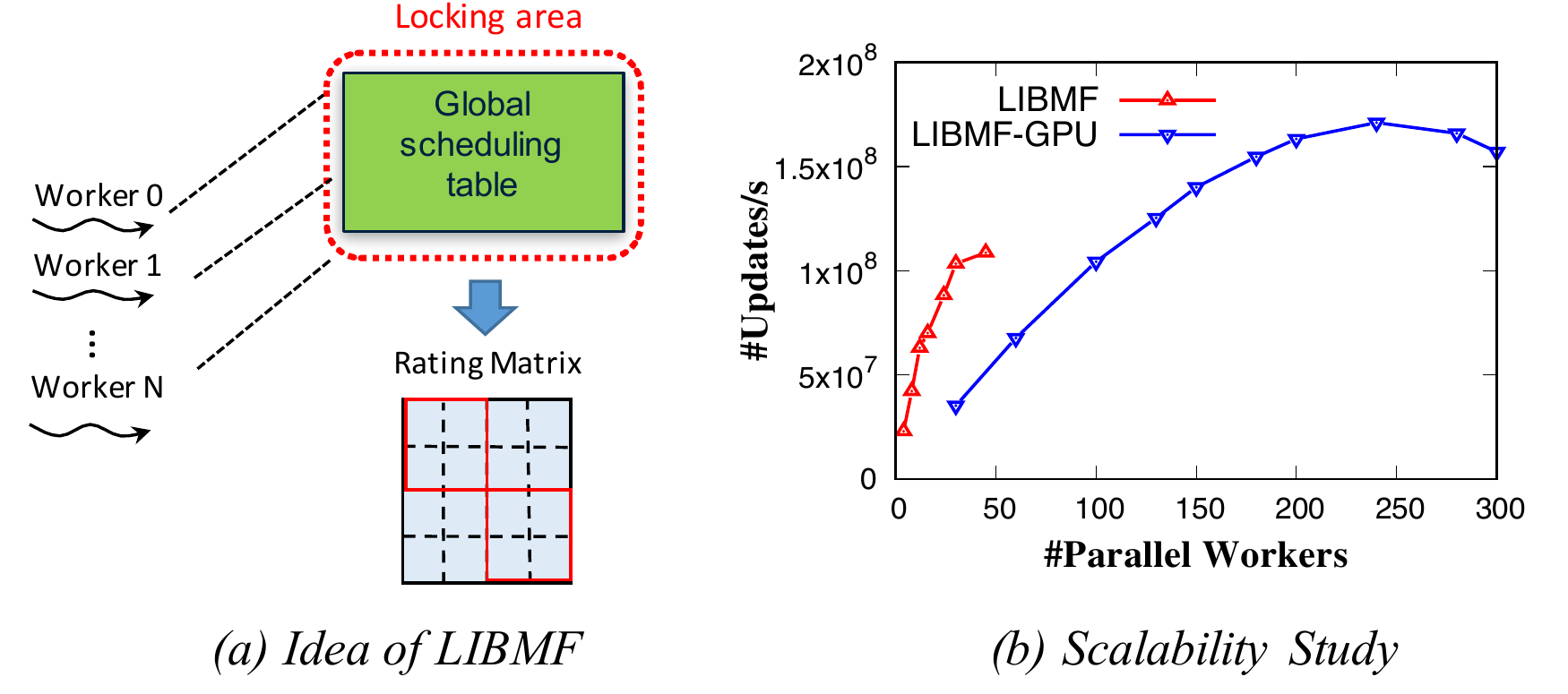}
\vspace{-0.1cm}
\caption{(a) LIBMF uses a centralized table to manage parallel workers. (b) LIBMF scales to only 30 CPU threads and 240 GPU thread blocks.}\label{figure:libmf}
\vspace{-0.1cm}
\end{figure}

The original SGD algorithm is serial, with samples in the rating matrix picked up randomly and updated in sequence. 
To exploit parallelism, a workload scheduling policy that assigns tasks to parallel workers becomes necessary. We start from investigating the existing CPU-based scheduling policies. Specifically, we select a representative system \textit{LIBMF}~\cite{tist15libmf}, a shared memory SGD solution to MF. LIBMF proposes a novel workload scheduling policy which successfully solves the load imbalance problem and achieve high efficiency. As shown in Figure~\ref{figure:libmf}(a), LIBMF divides the rating matrix to several blocks and uses a global scheduling table to manage the parallel workers. Whenever a worker is free, an idle independent block is scheduled to it. The process is repeated until convergence. However, we and others~\cite{ipdpsw15japan} observe that \textbf{LIBMF faces scalability issues because of the global scheduling table it uses}. 

Figure~\ref{figure:libmf}(b) shows a scalability study of LIBMF. \textit{LIBMF-GPU} is a GPU version of LIBMF that combines the workload scheduling policy of LIBMF and our GPU computation implementation described in Section~\ref{sec:vec}. We use SGD \textit{updates per second} as the performance metric:
\begingroup\makeatletter\def\f@size{8}\check@mathfonts
\begin{equation}
   \#Updates/s =  \frac{ \#Iterations \times \#Samples }{ Elapsed\ Time }
   \nonumber
\end{equation}
\endgroup
where $\#Iterations$, $\#Samples$, $Elapsed\ Time$ indicate number of iterations, number of non-zero samples in the input matrix $R$, and elapsed time in seconds, respectively. 

Evaluations show that the performance of LIBMF saturates around 30 concurrent workers (CPU threads), which is consistent with the previous study~\cite{ipdpsw15japan}. We perform some GPU-specific optimization techniques when implementing LIBMF-GPU,  however, it still can only scale to 240 thread blocks, much lower than the hardware limit(768 thread blocks). The reason why LIBMF cannot scale to many parallel workers is that, it uses a global scheduling table to manage all parallel workers. 

At each time, only one parallel worker can access the table and it is also time consuming to find a free block to assign the work to. Therefore, when the number of workers increase, the waiting time also increases. As the number of worker grows, the waiting time becomes dominating. This shows that, cuMF\_SGD can not simply re-use existing scheduling policies. To overcome this scheduling overhead, we propose two GPU-specific scheduling schemes, \textit{batch-Hogwild!} and \textit{Wavefront-update}. Batch-Hogwild! avoids block-based scheduling and improves the cache efficiency by process samples in batch. Wavefront-update is still block-based, but only requires a local look-up instead of the expensive global look-up in LIBMF.

\vspace{-0.05cm}
\vspace{-0.2cm}

\subsubsection{Batch-Hogwild!}\label{sec:batch-hogwild}
We propose batch-Hogwild!, a variant of Hogwild!~\cite{recht2011hogwild} with better cache efficiency. Hogwild! is efficient as its lock-free scheme incurs low scheduling overhead. It is not efficient, however, in terms of data locality~\cite{tist15libmf}. In vanilla Hogwild!, each parallel worker randomly selects one sample from the rating matrix at each step. After each update, Hogwild! may not access the consecutive samples in the rating matrix and corresponding rows and columns in the feature matrices during a long time interval, leading to low cache efficiency. As discussed in Section~\ref{sec:vec}, we carefully align the memory access to feature matrices to achieve perfect memory coalescing and the high memory bandwidth on GPUs makes accessing feature matrices no longer a performance bottleneck. To accelerate the access to rating matrix, we exploit the spatial data locality using L1 data cache. We let each parallel worker, instead of fetch one sample randomly at a time, fetches $f$ consecutive samples and update them serially. Note that these samples are consecutive in terms of their memory storage; because we shuffle samples, they are still random in terms of their coordinates in $R$. By doing so, the data locality is fully exploited. Consider the L1 cache line size is 128 bytes and the size of each sample is 12 bytes  (one floating point and two integers), $ f > 128/12$ is enough to exploit the locality. We evaluate different values of $f$ and find that they yield similar benefit. Therefore we choose $f=256$ without loss of generality.

\begin{figure}[h]
\centering
\includegraphics[scale=0.47,angle=0]{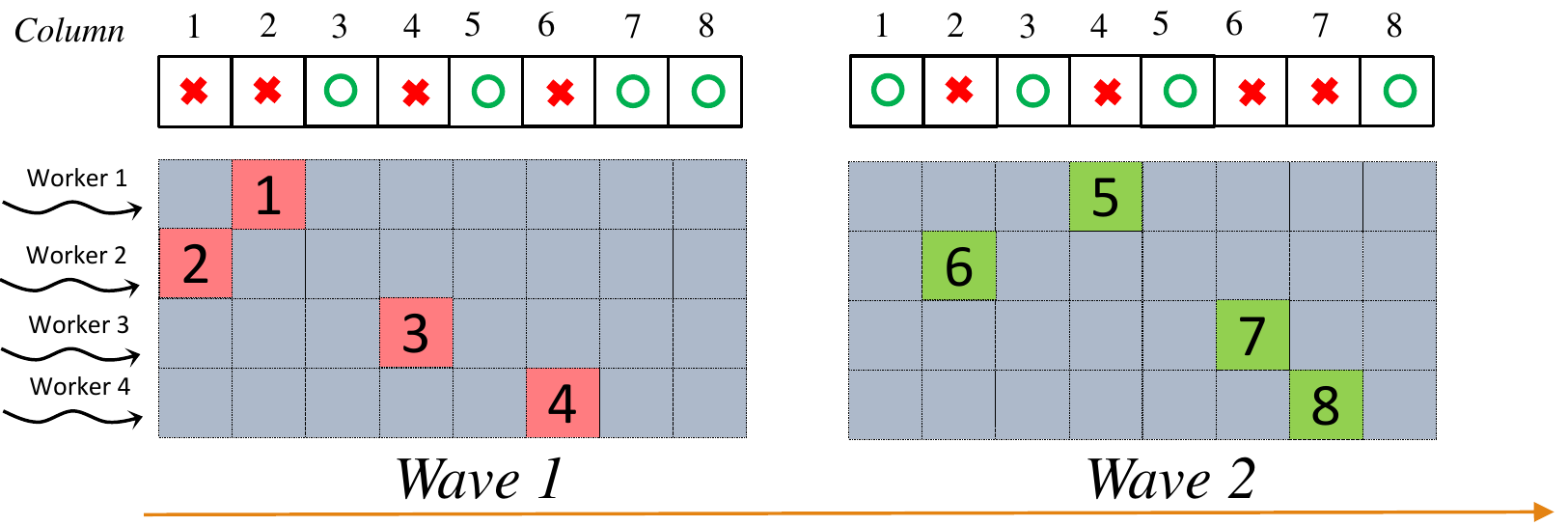}
\vspace{-0.2cm}
\caption{Wavefront-update. Each parallel worker is assigned to a row and pre-randomize its column update sequence. E.g., when Worker 3 completes Block 3 in Wave 1, it releases Column 4 such that Worker 1 can start Block 5 in Wave 2.} \label{figure:schedule}
\end{figure}
\vspace{-0.2cm}

\subsubsection{Wavefront-update}
As discussed, existing scheduling schemes \cite{tist15libmf, sigkdd11gemulla} impose a global synchronization, where all workers look up a global table to \textbf{find both row and column coordinates} to update. This is expensive and has been shown not scalable to the hundreds of workers on GPUs. To overcome this, we propose wavefront-update, a light-weight scheme that \textbf{locks and look up columns only}.

We explain the idea of wavefront-update using Figure~\ref{figure:schedule}. We use four parallel works to process an $R$ which is partitioned into $4 \times 8$ blocks. Each worker is assigned to a row in this $4 \times 8$ grid, and each generates a permutation of $\{1,2,3,...,7,8\}$ as its column update sequence. By this means, an epoch is conducted in eight waves given this sequence. 
In each wave, one worker update one block, and workers do not update blocks in the same column. Assume \textit{Worker 1} has the sequence defined as $\{2,4,...\}$ and \textit{Workder 3} has sequence $\{4,6,...\}$. With this sequence, \textit{Worker 1} updates \textit{Block 1} in wave 1 and \textit{Block 5} in wave 2. To avoid conflicts, we propose a light-weight synchronization scheme between waves using the column lock array. As shown the figure, we use an array to indicate the status of each column. Before a worker moves to next wave, it checks the status of the next column defined in its sequence. For example, after \textit{Worker 1} finishes \textit{Block 1}, it needs to check the status of column 4 and does not need to care about other columns' status. When \textit{Work 3} finishes \textit{Block 3} and releases column 4, \textit{Worker 1} is allowed to move to wave 2. 
There are two main benefits by doing so: (1) reduce the two-dimension look-up table in \cite{tist15libmf, sigkdd11gemulla} to an one-dimension array, (2) minimize the workload imbalance problem, as a worker can start the next block earlier compared to waiting for all other workers to finish. 

\begin{figure}[h]
\centering
\includegraphics[scale=0.35,angle=0]{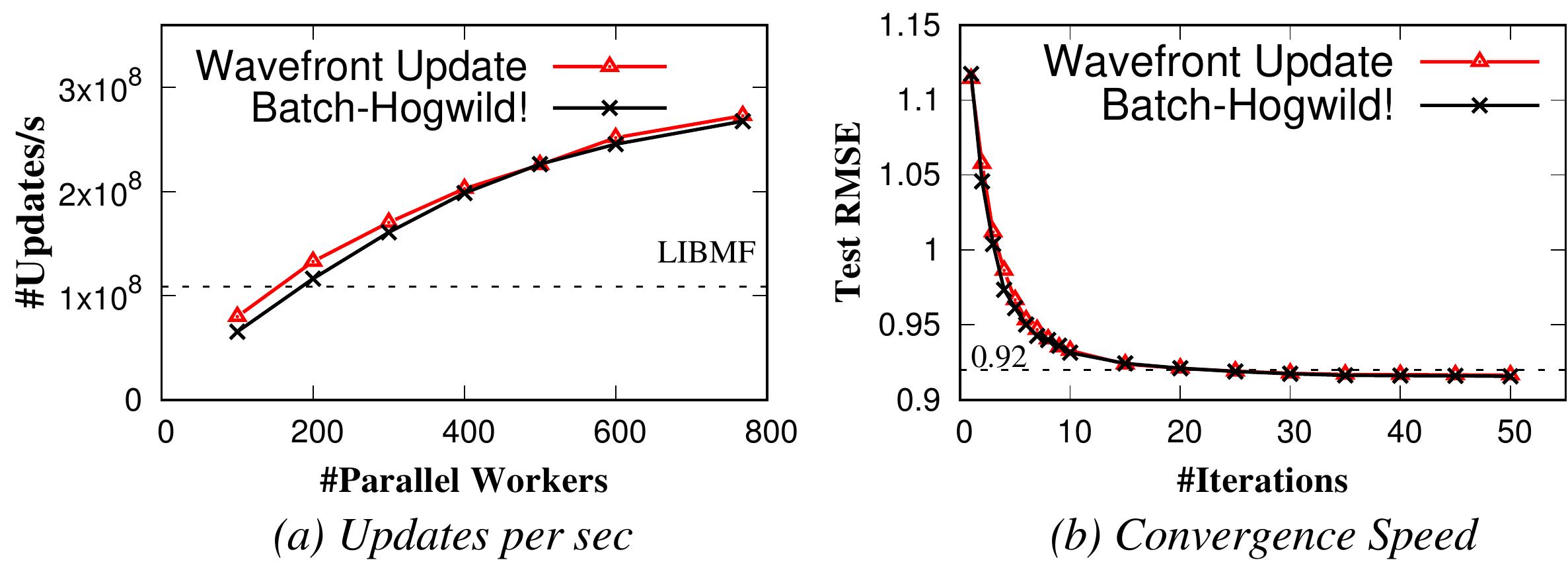}
\vspace{-0.2cm}
\caption{Performance Comparison of \textit{batch-Hogwild!} and \textit{wavefront-update} on \textit{Netflix} data set. Both techniques scale much better than LIBMF.}\label{figure:hogwild}
\vspace{-0.3cm}
\end{figure}

\subsubsection{Evaluation of scheduling schemes}
We evaluate both techniques in terms of performance and convergence speed using the \textit{Netflix} data set and the Maxwell platform\footnote{\scriptsize{Details of the data set and platform are presented in Section~\ref{sec:exp}.}}. We use metric $\#Updates/s$ to quantitative the performance. Figure~\ref{figure:hogwild}(a) shows the scalability of \textit{batch-Hogwild!} and \textit{Wavefront-update} with different number of parallel workers (i.e., thread blocks). When increasing the number of parallel workers, both techniques achieve near-linear scalability. When the number of parallel workers hits the hardware limit (768) of the Maxwell GPU, both techniques achieve $\scriptsize{\sim}$0.27 billion updates per second, which is 2.5 times of LIBMF. Therefore, we conclude that our proposed solutions can perfectly solve the scalability problem of the scheduling policy and fully exploit the equipped hardware resources on GPUs. We also evaluate the convergence speed of both techniques. We use the \textit{root mean square root} error on the standard test data set as the indication of convergence. Figure~\ref{figure:hogwild}(b) shows the decrease of \textit{Test RMSE} in iterations. Overall, batch-Hogwild! converges a little bit faster than Wavefront-update. The reason is, batch-Hogwild! enforces more randomness in update sequence, compared with the block-based wavefront-update. Based on this observation, we use batch-Hogwild! as the default scheme on one GPU.

\begin{figure*}
\centering
\includegraphics[scale=0.53,angle=0]{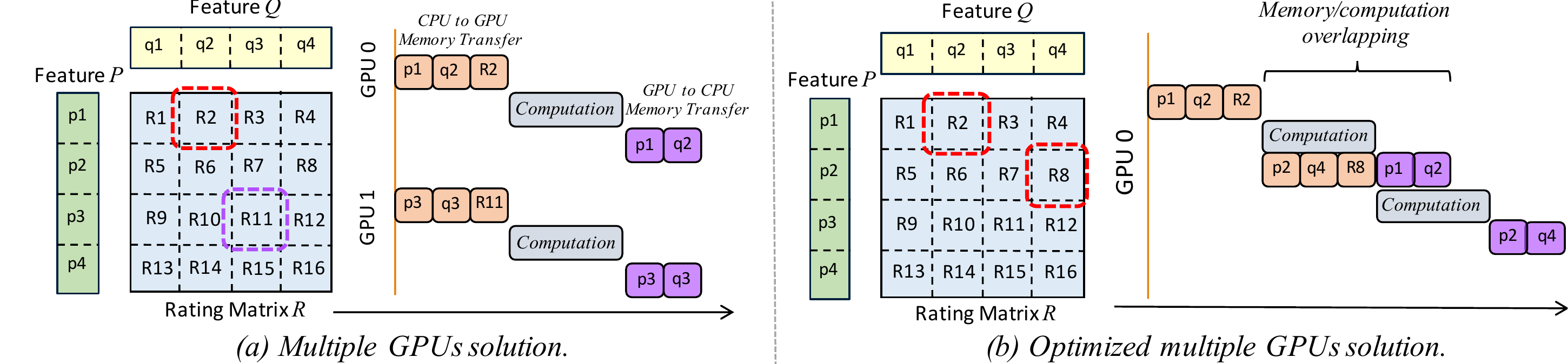}
\vspace{-0.3cm}
\caption{(a) Multiple GPUs solution of cuMF\_SGD, where the rating matrix is partitioned and each partition can fit into a GPU's device memory. (b) Optimizing the multi-GPU solution by overlapping memory transfer with computation. }\label{figure:multi}
\vspace{-0.38cm}
\end{figure*}

\section{Scale to Large Data sets}\label{sec:scale}

Section~\ref{sec:single} presents how to solve MF in a single GPU, assuming the rating matrix and feature matrices fully reside in GPU memory. However, the limited GPU memory capacity~\cite{ppopp08cuda} prevents cuMF\_SGD from solving large scale problems. For example, NVIDIA TITAN X GPU has 12 GB device memory that can only store 1 billion samples (one sample needs one float and two integers). Nowadays, real-world problems may have $10^{11}$ samples~\cite{hpdc16wei}. Techniques such as \textit{Unified Virtual Memory}~\cite{guide} allow GPU to use CPU memory but with high overhead. Consider these factors, to solve large-scale MF problems that can not fit into one GPU's device memory, we need to partition the data sets and stage the partitions to GPUs in batches. Moreover, We should overlap data transfer with computation to alleviate the delay caused by slow CPU-GPU memory transfer. Please note that, the partitions can be processed by one or multiple GPUs.

\subsection{Partition to Multiple GPUs}\label{sec:mgpus}

Figure~\ref{figure:multi} shows our proposed multiple GPUs solution. The main idea is to divide the rating matrix $R$ into multiple blocks; each block is small enough to fit into a GPU's device memory such that independent blocks can update concurrently on different GPUs. The multiple GPU solution works as follows,
\begin{enumerate}
\item Divide the rating matrix $R$ into $i \times j$ blocks. Meanwhile, divide feature matrix $p$ into $i$ segments and feature matrix $q$ into $j$ segments accordingly. 
\vspace{-0.15cm}
\item When a GPU is idle, randomly select one matrix block from those independent blocks and dispatch it to the GPU.
\vspace{-0.15cm}
\item Transfer the matrix block and corresponding feature sub-matrices $p$ and $q$ to the GPU. Then update the matrix block using the single GPU implementation discussed in Section~\ref{sec:single}. After the update, transfer $p$ and $q$ back to CPU.
\vspace{-0.15cm}
\item Iterate from 2 until convergence or the given number of iterations is reached.
\end{enumerate}

We further explain the proposed scheme using the example shown in Figure~\ref{figure:multi}(a). In Step 1, we divide $R$ into $4 \times 4$ blocks and use two GPUs. In Step 2, we send block $R2$ to GPU 0 and $R11$ to GPU 1. Again, consider the nature of MF, updating $R2$ only touches sub-matrices $p1$ \& $q2$ while updating $R11$ only touches $p3$ \& $q3$. Hence, GPU 0 only needs to store $R2$, $p1$, and $q2$ in its device memory while GPU 1 only needs to store $R11$, $p3$, and $q3$. By doing so, the problem is divided and conquered by multiple GPUs. After deciding the block scheduling order, cuMF\_SGD transfers $p1$, $q2$, $R2$ to GPU 0 and $p3$, $q3$, $R11$ to GPU 1, then performs the computation on two GPUs in parallel. The GPU side computation follows the rules we discussed in Section~\ref{sec:single}. After finishing the computation, the updated $p1$, $q2$, $p3$, and $q3$ are transferred back to CPU memory. Note that we don't have to transfer $R2$ or $R11$ back to CPU memory as they are read-only. 

\textbf{Scalability problem}. We mentioned LIBMF faces scalability issue, as the scheduling overhead increases quickly with the number of workers~\cite{ipdpsw15japan}. Our multiple-GPU scheduling scheme has similar complexity with that of LIBMF. However, it does not face the same scalability issue as we only need to schedule to a few GPUs instead of hundreds of workers.

\subsection{Overlap Data Transfer and Compute} \label{sec:optimization}

GPUs' memory bandwidth are much larger than the CPU-GPU memory transfer bandwidth. For example, NVIDIA TITAN X GPUs provide 360 GB/s device memory bandwidth while the CPU-GPU memory bandwidth is only $\scriptsize{\sim}$16 GB/s (PCIe v3 16x). In single GPU implementation, CPU-GPU memory transfer only happens at the start and end of MF, and therefore not dominant. However, when the data set can not fit into the GPU memory, memory transfer happens frequently and has higher impact on the overall performance. 

Given the memory transfer overhead, we overlap the memory transfers and computation when solving large problems, as shown in Figure~\ref{figure:multi}(b). Due to space limitation , we only plot one GPU. The key idea is, at the block scheduling time, instead of randomly selecting one independent block for the GPU, the optimized technique randomly \textbf{selects multiple blocks at a time}. Those blocks are pipelined to overlap the memory transfer and computation: we schedule two blocks to GPU 0, and overlap the memory transfer of the second block ($R8$) with the computation of the first block ($R2$). Note that the two blocks scheduled to one GPU do not need to be independent as they are updated in serial; meanwhile, blocks scheduled to different GPUs have to be independent with each other to avoid conflicts. By doing so, we can reduce the overhead of slow CPU-GPU memory transfer and improve the overall performance.

\textbf{Discussion}. Allocating more blocks to one GPU would yield more performance benefit as more memory/computation overlapping can be achieved. However, the number of available blocks is limited by how do we divide the rating matrix $R$. Consider we divide $R$ to $i \times i$ and we have two GPUs running in parallel, the number of blocks per GPU cannot be more than $i/2$. In practice, $i$ is determined by the size of the rating matrix $R$ and the available hardware resources on the GPU. We will discuss it in Section~\ref{sec:convergence}.

\vspace{-0.15cm}
\subsection{Implementation Details}
\textbf{Multiple GPUs management}. We implement it using multiple CPU threads within one process. Within the process, there is one \textit{host thread} and multiple \textit{worker threads}, where each GPU is bound to one worker thread. The host thread manages the workload scheduling and informs worker threads of the scheduling decision. Each worker thread then starts data transfer and launches compute kernels on a GPU.

\textbf{Overlapping}. Each worker thread is responsible to overlap the computation and CPU-GPU memory transfers. We use CUDA \textit{streams} to achieve this. A \textit{stream} contains a list of GPU commands that are executed in serial, and commands in different streams are executed in parallel if hardware resources permit. Each worker thread uses three streams that manage CPU-GPU memory transfer, GPU-CPU memory transfer, and GPU kernel launch, respectively.

\section{Experiments}\label{sec:exp}

We implement cuMF\_SGD using CUDA C (source code at \url{http://github.com/cumf/cumf_sgd/}), evaluate its performance on public data sets, and demonstrate its advantage in terms of performance and cost. Section~\ref{sec:setup} introduces the experimental environment. The following experiments are designed to answer these questions:
\begin{itemize}
\item Compared with state-of-the-art SGD-based approaches on CPUs~\cite{libmf++,vldb14nomad}, is cuMF\_SGD better and why? (Section~\ref{sec:evaluation})
\vspace{-0.1cm}
\item What is the implication of using different generations of GPUs? (Section~\ref{sec:gpuarch})
\vspace{-0.1cm}
\item Compared with the ALS-based GPU library \textbf{cuMF\_ALS} that we published earlier~\cite{hpdc16wei}, what is the advantage of cuMF\_SGD? (Section~\ref{sec:ALS})
\vspace{-0.1cm}
\item Parallelizing SGD is always tricky and may lead to converge problems. What are the factors impacting parallelizing SGD? (Section~\ref{sec:convergence})
\vspace{-0.1cm}
\item When scale up to multiple GPUs, is cuMF\_SGD still efficient? (Section~\ref{sec:multiple})
\end{itemize}

\subsection{Experimental Setup}\label{sec:setup}
\textbf{Platform}. We evaluate cuMF\_SGD on heterogeneous platforms with both CPU and GPUs. Table~\ref{table:platform} shows the configuration of the two servers used in experiments. 
\vspace{-0.4cm}
\begin{table}[h]
\scriptsize
    \begin{center}
      \caption{Configuration of the Maxwell \protect\cite{maxwell} and Pascal \protect\cite{pascal} Platform.}
    \begin{tabular}{| >{\centering\arraybackslash}m{0.7 cm} | >{\centering\arraybackslash}m{7 cm} |}
    \hline
     \multicolumn{2}{|c|}{ \textbf{Maxwell Platform}} \\ \hline
     CPU & \vspace{0.06cm} 12-core Intel Xeon CPU E5-2670*2 (up to 48 threads), 512 GB memory \\[0ex] \hline
     GPU & \vspace{0.06cm} TITAN X GPU*4, per GPU: 24 SMs, up to 768 thread blocks, 12 GB device memory. \\[0ex] \hline
	\multicolumn{2}{|c|}{\textbf{Pascal Platform}} \\ \hline
	 CPU & \vspace{0.06cm} 2*10 PowerNV 8 processors with SMT 8 and NVLink.\\[0ex] \hline
     GPU & \vspace{0.06cm} Tesla P100 GPU*4, per GPU: 56 SMs, up to 1792 thread blocks, 16 GB device memory. \\[0ex] \hline
  \end{tabular}
  \label{table:platform}
  \end{center}
  \vspace{-0.4cm}
\end{table}

\textbf{Data sets}. We use three public data sets: \textit{Netflix}, \textit{Yahoo!Music}, and \textit{Hugewiki}. Details of them are shown in Table~\ref{table:dataset}. \textit{Netflix} and \textit{Yahoo!Music} come with a test set but \textit{Hugewiki} does not. We randomly sample and extract out 1\% of the data set for testing purpose.

 \vspace{-0.4cm}
\begin{table}[h]
\scriptsize
   \begin{center}
      \caption{Details of data sets used.}
    \begin{tabular}{|c |c |c |c|}
    \hline
     \textbf{Dataset}  & \textit{Netflix} & \textit{Yahoo!Music} & \textit{Hugewiki}       \\ \hline
      $m$              & 480,190          & 1,000,990            & 50,082,604              \\ \hline
      $n$              & 17,771           & 624,961              & 39,781                  \\ \hline
      $k$              & 128              & 128                  & 128                     \\ \hline
      $Train\ Set$     & 99,072,112       & 252,800,275          & 3,069,817,980           \\ \hline
      $Test\ Set$      & 1,408,395        & 4,003,960            & 31,327,899              \\ \hline
  \end{tabular}
  \label{table:dataset}
  \end{center}
   \vspace{-0.2cm}
\end{table}

\vspace{-0.1cm}
\textbf{Parameter}. As mentioned in the introduction, this paper focus on system-level but not algorithmic-level optimization. Therefore, we did not spend much effort in parameter turning. Instead, we use the parameters adopted by earlier work~\cite{hpdc16wei,vldb14nomad,tist15libmf,libmf++}. For learning rate, we adopt the learning rate scheduling techniques used by Yun et al.~\cite{vldb14nomad}, where the learning rate $s_t$ at epoch $t$ is monotonically reduced in the following routine:
\begingroup\makeatletter\def\f@size{8}\check@mathfonts
\begin{equation}
   s_t = \frac{ \alpha }{1 + \beta \cdot t^{1.5} }
   \nonumber
\end{equation}
\endgroup
$\alpha$ is the given initial learning rate and $\beta$ is another given parameter. The parameters used by cuMF\_SGD are listed in Table~\ref{table:parameter}.

\begin{table}[h]
    \begin{center}
      \caption{Used parameters for each dataset.}
    \begin{tabular}{| >{\centering\arraybackslash}p{2.5cm} | >{\centering\arraybackslash}p{0.8cm}| >{\centering\arraybackslash}p{0.8cm} | >{\centering\arraybackslash}p{0.8cm} |}
    \hline
     \textbf{Dataset}     & $\lambda$ & $\alpha$ & $\beta$  \\ \hline
     \textit{Netflix}     & 0.05      & 0.08     & 0.3      \\ \hline
     \textit{Yahoo!Music} & 1.0       & 0.08     & 0.2      \\ \hline
     \textit{Hugewiki}    & 0.03      & 0.08     & 0.3      \\ \hline
  \end{tabular}
  \label{table:parameter}
  \end{center}
\end{table}

\subsection{Comparison of SGD approaches}\label{sec:evaluation}
We compare cuMF\_SGD with the following state-of-the-art approaches.
\begin{itemize}
\item \textbf{LIBMF}~\cite{tist15libmf}. LIBMF is a representative blocking-based solution on shared-memory systems. Its main design purpose is to balance the workload across CPU threads and accelerate the memory access. It leverages SSE instructions and a novel learning rate schedule to speed up the convergence~\cite{libmf++}. 

We exhaustively evaluate all possible parallel parameters on the Maxwell platform and select the optimal one. For example, we use 40 CPU threads and divide the input rating matrix $R$ into $100 \times 100$ blocks; we set its initial learning rate as $0.1$. 

\item \textbf{NOMAD}~\cite{vldb14nomad}. NOMAD is a representative distributed matrix factorization solution. It uses a 64-node HPC cluster to solve MF. It proposes a decentralized scheduling policy to reduce the synchronization overhead and discusses how to reduce the inter-node communication. We present the best results presented in the original paper, i.e., using 32 nodes for \textit{Netflix} and \textit{Yahoo!Music} data sets and using all 64 nodes for \textit{Hugewiki} data set on the HPC cluster.


\begin{figure*}
\centering
\includegraphics[scale=0.33,angle=0]{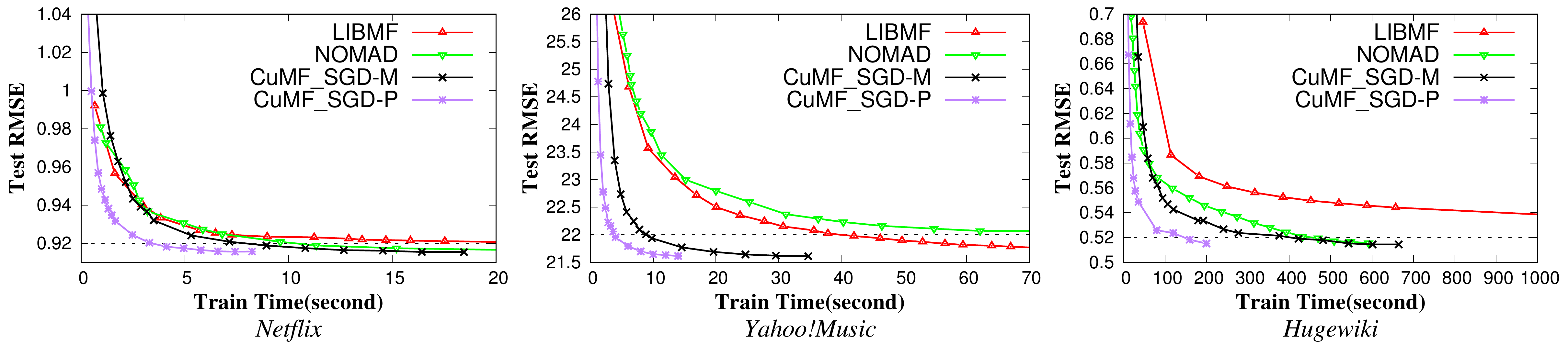}
\vspace{-0.2cm}
\caption{Test RMSE over training time on three data sets. CuMF\_SGD converges faster than all other approaches with only one GPU card.}\label{figure:compare}
\vspace{-0.3cm}
\end{figure*}
\item \textbf{CuMF\_SGD}. We evaluate cuMF\_SGD on both Maxwell and Pascal platforms, with all three data sets. We name the results on Maxwell as \textit{cuMF\_SGD-M} and those on Pascal as \textit{cuMF\_SGD-P}. We use one GPU in this subsection. The number of parallel workers (thread blocks) is set as the maximum of the corresponding GPU architecture (768 on Maxwell platform and 1792 on Pascal platform). As \textit{Hugewiki} can not fit into one GPU's memory, we divide it into $64 \times 1$ blocks and at each scheduling, we schedule 8 blocks to overlap memory transfer and computation.

\end{itemize}

Figure~\ref{figure:compare} shows the test RMSE w.r.t. the training time. Table~\ref{table:time} summarizes the training time required to converge to a reasonable RMSE (0.92, 22.0, and 0.52 for \textit{Netflix}, \textit{Yahoo!Music}, and \textit{Hugewki}, respectively). Results show that \textbf{with only one GPU, cuMF\_SGD-P and cuMF\_SGD-M perform much better (3.1X to 28.2X) on all data sets compared to all competitors}, including NOMAD on a 64-node HPC cluster. In the following we analyze the reasons.

\begin{table}[h]
\scriptsize
    \begin{center}
      \caption{Training time speedup normalized by LIBMF.}
    \begin{tabular}{| >{\centering\arraybackslash}p{2.1cm} | >{\centering\arraybackslash}p{1.22cm} |>{\centering\arraybackslash}p{1.8cm}| >{\centering\arraybackslash}p{1.65cm} |}
    \hline
     \textbf{Data set}     & \textit{Netflix} & \textit{Yahoo!Music} & \textit{Hugewiki}  \\ \hline
     {LIBMF}       & 23.0s            & 37.9s                & 3020.7s            \\ \hline
     {NOMAD}    & 9.6s(\scriptsize{2.4X})             & 108.7s(\scriptsize{0.35X})                 & 459.1s(\scriptsize{6.6X})             \\ \hline
     {CuMF\_SGD-M} & 7.5s(\scriptsize{3.1X})             & 8.8s(\scriptsize{4.3X})                 &442.3s(\scriptsize{6.8X})                \\ \hline
     {CuMF\_SGD-P} & 3.3s(\scriptsize{7.0X})             & 3.8s(\scriptsize{10.0X})                 & 107.0s(\scriptsize{28.2X})             \\ \hline
  \end{tabular}
  \label{table:time}
  \end{center}
  \vspace{-0.2cm}
\end{table}

\begin{figure}[h]
\centering
\includegraphics[scale=0.26,angle=0]{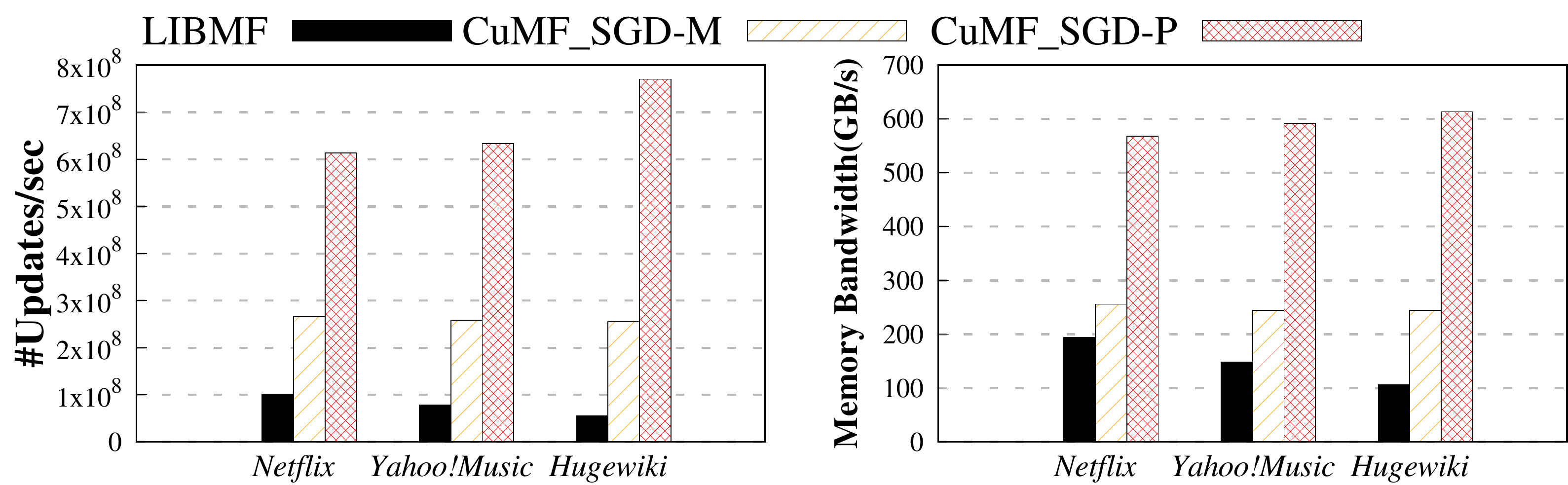}
\caption{Updates per second and achieved memory bandwidth of LIBMF, cuMF\_SGD-M, and cuMF\_SGD-P. The achieved memory bandwidth explains the advantage of cuMF\_SGD.}
\label{figure:bandwidth}
\vspace{-0.3cm}
\end{figure}

\textbf{Comparison with LIBMF}. As shown in Figure~\ref{figure:compare} and Table~\ref{table:time}, cuMF\_SGD outperforms LIBMF on all data sets, on both Maxwell and Pascal. More precisely, cuMF\_SGD-M is 3.1X - 6.8X as fast as LIBMF and cuMF\_SGD-P is 7.0X - 28.2X as fast. CuMF\_SGD outperforms LIBMF because it can do more updates per second, as shown in Figure~\ref{figure:bandwidth}(a). We already mentioned that that matrix factorization is memory bound; LIBMF is also aware of that and strives to keep all frequently used data in the CPU cache. However, the limited cache capacity on a single CPU makes LIMBF suboptimal in large data sets. As shown in Figure~\ref{figure:bandwidth}(b), LIBMF achieves an effective memory bandwidth of 194 GB/s\footnote{\scriptsize{The achieved memory bandwidth measures the data processed by the compute units per second, and can be higher than the theoretical off-chip memory bandwidth thanks to the cache effect.}} on the \textit{Netflix} data set (with 99M samples) -- close to cuMF\_SGD-M. However its achieved bandwidth drops almost by half, to 106 GB/s on the larger \textit{Hugewiki} data set (with 3.1B samples) -- while cuMF\_SGD achieves similar bandwidth in all data sets. 

On the scheduling policy of LIBMF. Simply porting LIBMF to GPUs leads to resource under-utilization due to the scalability of it scheduling policy (recall Figure~\ref{figure:libmf}). In contrast, the workload scheduling policy and memory/computation pattern of cuMF\_SGD are specifically designed to fully exploit the computation and memory resources on GPUs. Hence, as shown in Figure~\ref{figure:bandwidth} (b), cuMF\_SGD achieves much higher bandwidth than LIBMF. Moreover, cuMF\_SGD uses half-precision (2 bytes for a float number) to store feature matrices. As a result, it can perform twice updates as LIBMF with the same bandwidth consumption. 

\textbf{Compared with NOMAD}. As presented in~\cite{vldb14nomad}, NOMAD uses 32 nodes for \textit{Netflix} and \textit{Yahoo!Music} and 64 HPC nodes for \textit{Hugewiki}. Despite of the tremendous hardware resources, NOMAD is still outperformed by cuMF\_SGD on all data sets. As observed in Section~\ref{sec:introduction}, MF is a memory bounded and data communication happens frequently between parallel workers. When NOMAD distributes parallel workers to different nodes, the network bandwidth which is much less than intra-node communication, becomes the bottleneck.  Consequently, NOMAD achieves suboptimal scalability when scale from single node to multiple nodes, especially for  small data sets. For example, on \textit{Yahoo!Music}, NOMAD performs even worse than LIBMF that uses only one machine. 

NOMAD (on a 64-node HPC cluster) has similar performance with cuMF\_SGD-M on \textit{Hugewiki}, while it is much slower than cuMF\_SGD-P. Obviously, cuMF\_SGD is not only faster, using a single CPU card, it is also more cost-efficient.

\begin{figure*}
\centering
\includegraphics[scale=0.33,angle=0]{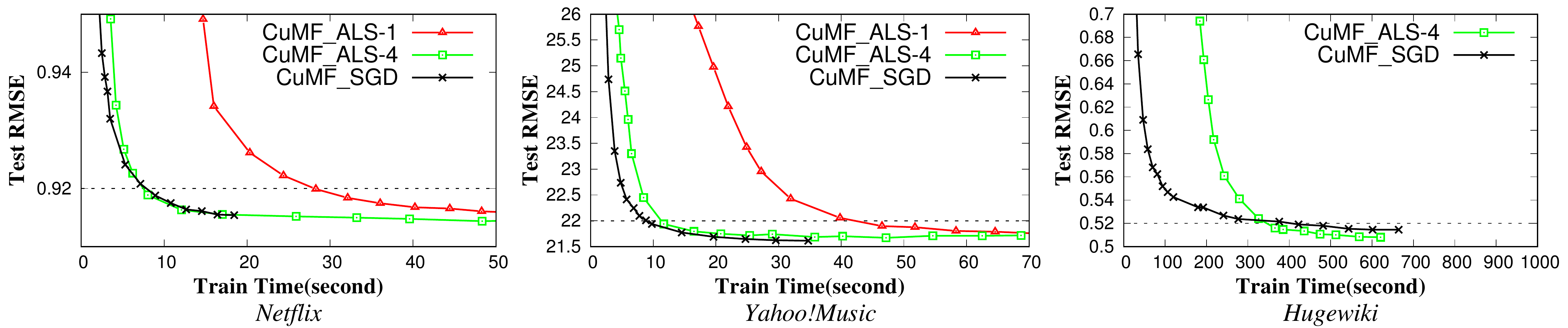}
\vspace{-0.3cm}
\caption{CuMF\_SGD vs. cuMF\_ALS. With one GPU, CuMF\_SGD converges faster than cuMF\_ALS-1 (one GPU) and similar to cuMF\_ALS-4 (four GPUs).}\label{figure:als}
\vspace{-0.2cm}
\end{figure*}

\begin{figure}[h]
\centering
\includegraphics[scale=0.26,angle=0]{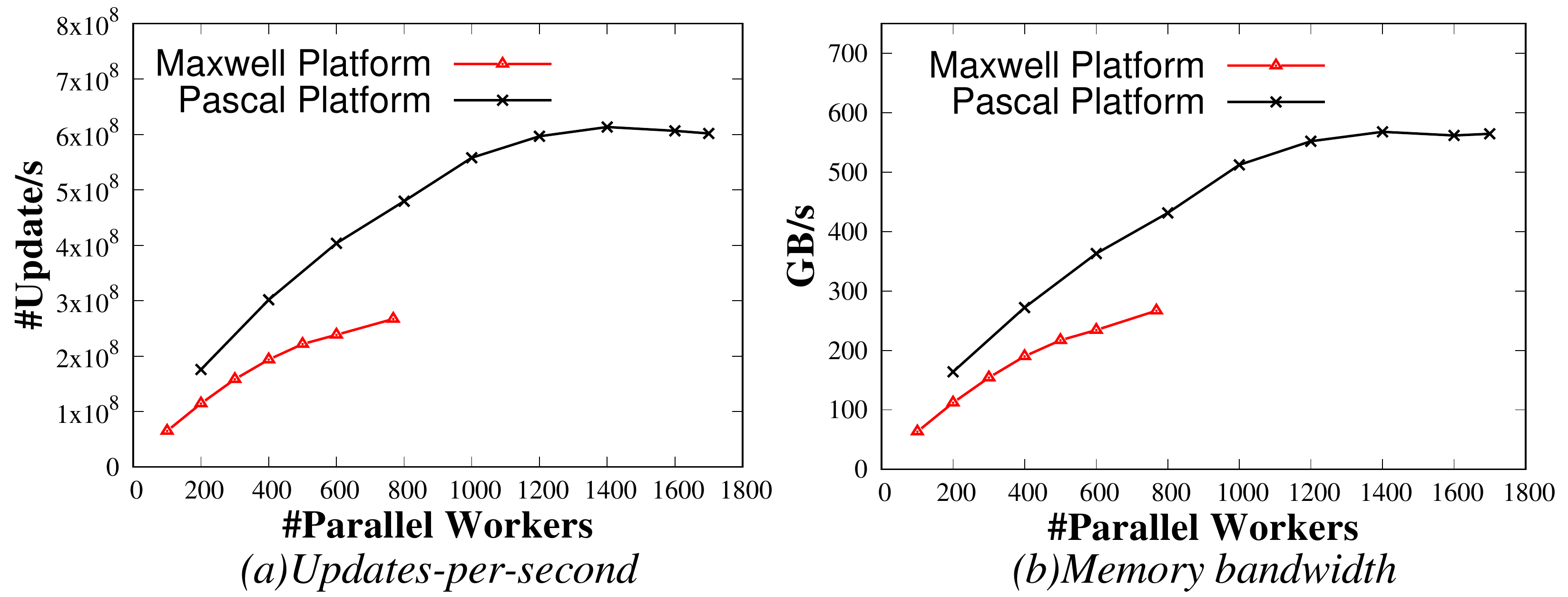}
\caption{Updates-per-second and achieved memory bandwidth cuMF\_SGD on Maxwell and Pascal, using the \textit{Netflix} data set. CuMF\_SGD performs better on the more recent Pascal platform. }\label{figure:cross}
\vspace{-0.2cm}
\end{figure}

\subsection{Implication of GPU Architectures}\label{sec:gpuarch}
We have evaluated cuMF\_SGD on the two current generations of GPUs: Maxwell and Pascal. We believe that cuMF\_SGD is able to scale to future GPU architectures with minor tuning effort. In this section, we explain the performance gap between Maxwell and Pascal in three aspects: computation resources, off-chip memory bandwidth, and CPU-GPU memory bandwidth. 

\textbf{Computation resources}. We show the SGD updates-per-second metric of two platforms with different number of parallel workers using \textit{Netflix} in Figure~\ref{figure:cross}(a). Results show that the Pascal platform scales to more parallel workers and achieves much higher $\#updates/s$ than Maxwell. This is because the Maxwell platform has 24 streaming multiprocessors (SMs) within each GPU, with each SM allowing up to 32 parallel workers (thread blocks). Hence, one Maxwell GPU allows up to 768 parallel workers. Meanwhile, the Pascal  GPU used has 56 SMs and allows 32 thread blocks on each SM. Hence, a Pascal GPU allows up to 1792 parallel workers, which is 2.3 times of that of Maxwell GPU. Overall, a Pascal GPU is more powerful than a Maxwell GPU in term of the amount of computation resources. 

\textbf{Off-chip memory bandwidth}. As we discussed before, SGD is memory bound. Optimized for throughput, GPUs are able to overlap memory access and computation by fast context switch among parallel workers~\cite{guide}. When there are enough parallel workers running on GPUs, long memory latencies can be hidden, which is exactly what happens with cuMF\_SGD. In this scenario, memory bandwidth, instead of memory latency, becomes the limitation of the performance. Pascal platforms provides twice as much theoretical peak off-chip memory bandwidth (780 GB/s) as Maxwell platforms(360 GB/s). Figure~\ref{figure:cross}(b) shows the achieved memory bandwidth on two platforms with different number of parallel workers. On Maxwell and Pascal, cuMF\_SGD achieves up to 266 GB/s and 567 GB/s, respectively. 

\textbf{CPU-GPU memory bandwidth}. \textit{Netflix} and \textit{Yahoo!Music} data sets are small enough to fit into the GPU device memory. For \textit{Hugewiki}, memory transfer occurs multiple times as the data cannot fit into GPU device memory. In Section~\ref{sec:optimization}, we propose to overlap data transfer with computation. Despite of this optimization, the CPU-GPU memory bandwidth still has noticeable impact on the overall performance as the perfect overlapping cannot be achieved. On the Maxwell platform, the memory transfer between CPU and GPU is via PCIe v3 16x with 16 GB/s bandwidth (we observe that on average, the achieved bandwidth is 5.5 GB/s). The very recent Pascal platform is with NVLink~\cite{nvlink} that can provide 40 GB/s in theory (we observe an average 29.1 GB/s CPU-GPU memory transfer bandwidth, which is 5.3X as that on Maxwell). This also explains why cuMF\_SGD achieves much more speedup on \textit{Hugewiki} using Pascal platform (28.2X) than that on Maxwell platform (6.8X).

\subsection{Comparison with cuMF\char`_ALS} \label{sec:ALS}

Our earlier work \textbf{cuMF\_ALS}~\cite{hpdc16wei} represents the state-of-art ALS-based matrix factorization solution on GPUs. We use one GPU for cuMF\_SGD, and one and four GPUs for cuMF\_ALS. Figure~\ref{figure:als} compares their performance on three data sets on Maxwell. We observe that cuMF\_SGD is faster than cuMF\_ALS-1 and achieves similar performance with cuMF\_ALS-4 with only one GPU.

It expected that cuMF\_SGD is faster than cuMF\_ALS, with the following reason. Each epoch of SGD needs memory access of $\mathcal {O}(N*k)$ and computation of $\mathcal {O}(N*k)$. Each epoch of ALS needs memory access of $\mathcal {O}(N*k)$ and computation of $\mathcal {O}(N*k^2+ (m+n)*k^3)$. ALS's epochs run slower due to its much more intensive computation. Although ALS needs fewer epochs to coverage, as a whole it converges slower. Despite the fact that cuMF\_ALS is slower than cuMF\_SGD, we still maintain both solutions at \url{https://github.com/cuMF/} because they serve different purposes: SGD converges fast and easy to do incremental update, while ALS is easy to parallelize and is able to deal with non-sparse rating matrices~\cite{computer09mf}.

\subsection{Convergence Analysis}\label{sec:convergence}
The original SGD algorithm is serial. To speed it up, we discuss how to parallelize it on one GPU in Section~\ref{sec:scheduling} and on multiple GPUs in Section~\ref{sec:mgpus}. It is well-known that SGD parallelization may have subtle implications on convergence~\cite{tist15libmf,vldb14nomad}. In the context of matrix factorization, the implication varies on the two schemes proposed in Section~\ref{sec:scheduling}: Hogwild! and matrix-blocking. 

\subsubsection{Hogwild!}
For one GPU, Section~\ref{sec:batch-hogwild} proposes the batch-Hogwild! scheme to partition work . As a vectorized version of Hogwild!, batch-Hogwild! inherits the limitation of Hogwild!. Given a rating matrix of $m \times n$ and $s$ parallel workers, convergence is ensured only when the following condition satisfied~\cite{recht2011hogwild}:
$$s \ll min(m, n) $$
For multiple GPUs, Section~\ref{sec:scale} proposes to first divide the rating matrix $R$ into $i \times  j$ blocks and process one block on one GPU in parallel if possible. In this case, the above condition needs to change to:
$$ s \ll min(\lfloor m/i \rfloor, \lfloor n/j \rfloor)$$
Our empirical study on \textit{Hugewiki} data set shows that, $s$ needs to be $< 1/20*min(\lfloor m/i \rfloor, \lfloor n/j \rfloor)$ to converge. We validated that, \textit{Hugewiki} data set has $min(m, n)=40k$ and we choose $s=768$; convergence is achieved when $j<40k/20/768\approx2$, and fails when $j=4$. We believe this is a limitation for all Hogwild!-style solutions.


\subsubsection{Matrix-blocking}
\begin{figure}[h]
\centering
\includegraphics[scale=0.26,angle=0]{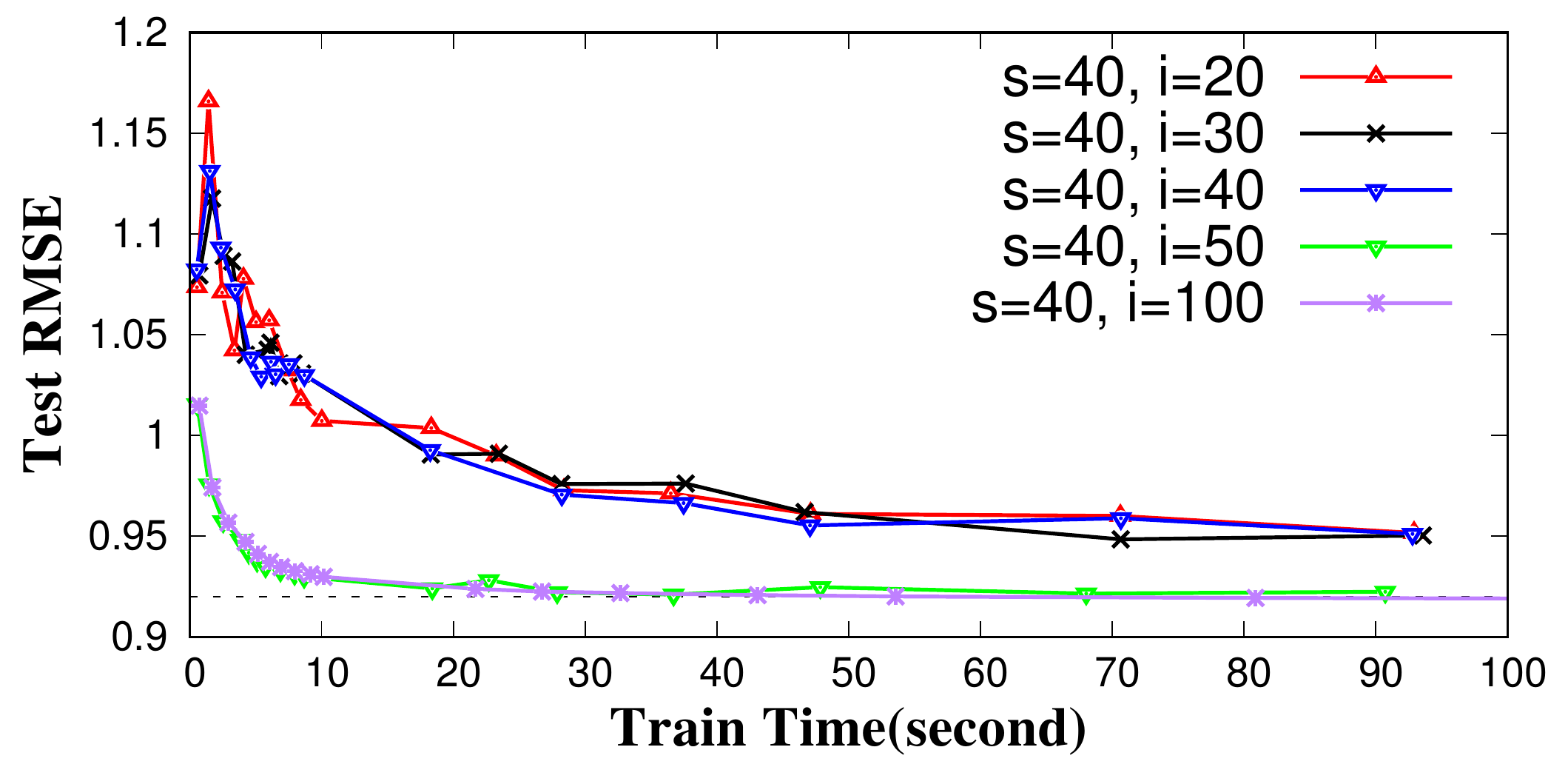}
 \vspace{-0.2cm}
\caption{Convergence speed of LIBMF on \textit{Netflix}. We fix \#parallel-workers $s=40$ and vary value $i$ to partition to $i\times i$ blocks.} \label{figure:convergence}
 \vspace{-0.5cm}
\end{figure}

\begin{figure}[h]
\centering
\includegraphics[scale=0.40,angle=0]{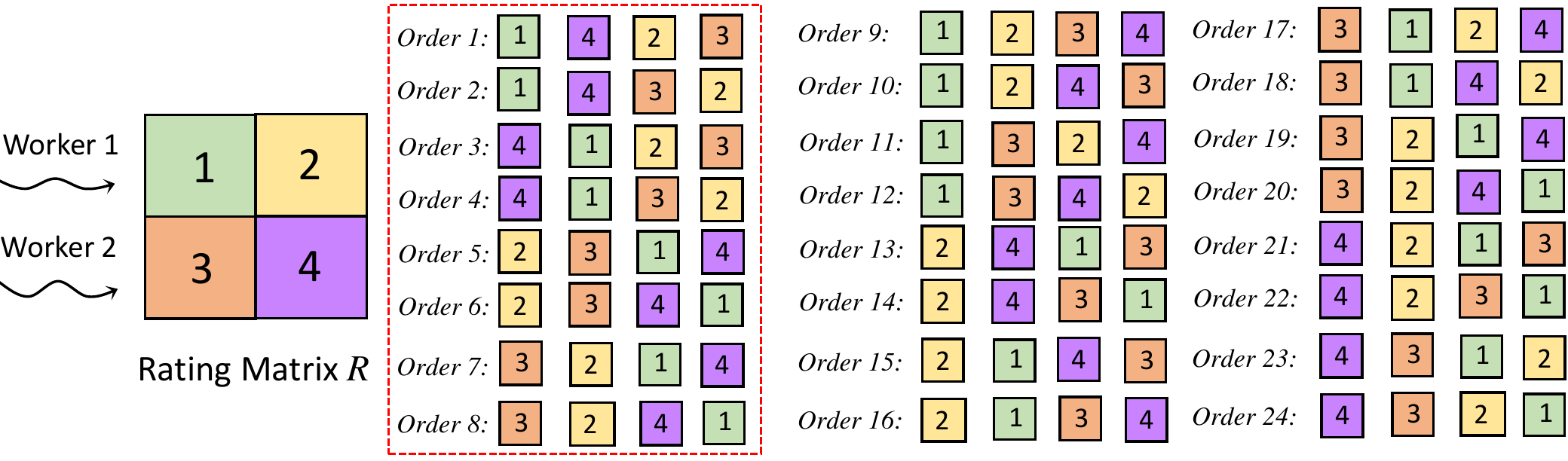}
 \vspace{-0.5cm}
\caption{A simple example to demonstrate the limitation of matrix blocking. The rating matrix is divided into $2 \times 2$ blocks and updated using two parallel workers.} \label{figure:blocking}
 \vspace{-0.4cm}
\end{figure}

The purpose of matrix-blocking is to avoid conflicts between parallel workers. However, we observe that matrix-blocking can have negative impact on convergence. Figure~\ref{figure:convergence} illustrates the the convergence speed of LIBMF on \textit{Netflix} data set with different parameters. In this study, we fix the number of parallel workers $s=40$; without loss of generality, we divide $R$ into $i \times i$ blocks and vary the value of $i$. Figure~\ref{figure:convergence} shows that when $i$ is less than or close to $s$, convergence speed is much slower or even cannot be achieved. We have similar observations on other data sets and using cuMF\_SGD. We briefly explain the reason with a simple example shown in Figure~\ref{figure:blocking}.

In Figure~\ref{figure:blocking}, we divide the rating matrix into $2 \times 2$ blocks and use 2 parallel workers. In theory, 4 blocks can have $4\times 3 \times 2 \times 1= 24$ possible update orders. We also show all update orders in Figure~\ref{figure:blocking}. However, only orders 1$\sim$8 out of the total 24 are feasible so as to avoid update conflicts. For example, when \textit{Block 1} is issued to one worker, only \textit{Block 4} can be issued to another worker. Hence, Blocks 2 and 3 cannot be updated between 1 and 4, which precludes order 9$\sim$12. This demonstrated that when $s \ge i$, all independent blocks have to be updated concurrently to make all workers busy, which enforces certain update order constraints and hurts the randomness. As a consequence, convergence speed can deteriorate. In practice, when cuMF\_SGD uses two GPUs, $R$ should at least be divided into $4 \times 4$ blocks.

 \vspace{-0.1cm}
\subsection{Scale Up to Multiple GPUs}\label{sec:multiple}

\begin{figure}[h]
\centering
\includegraphics[scale=0.31,angle=0]{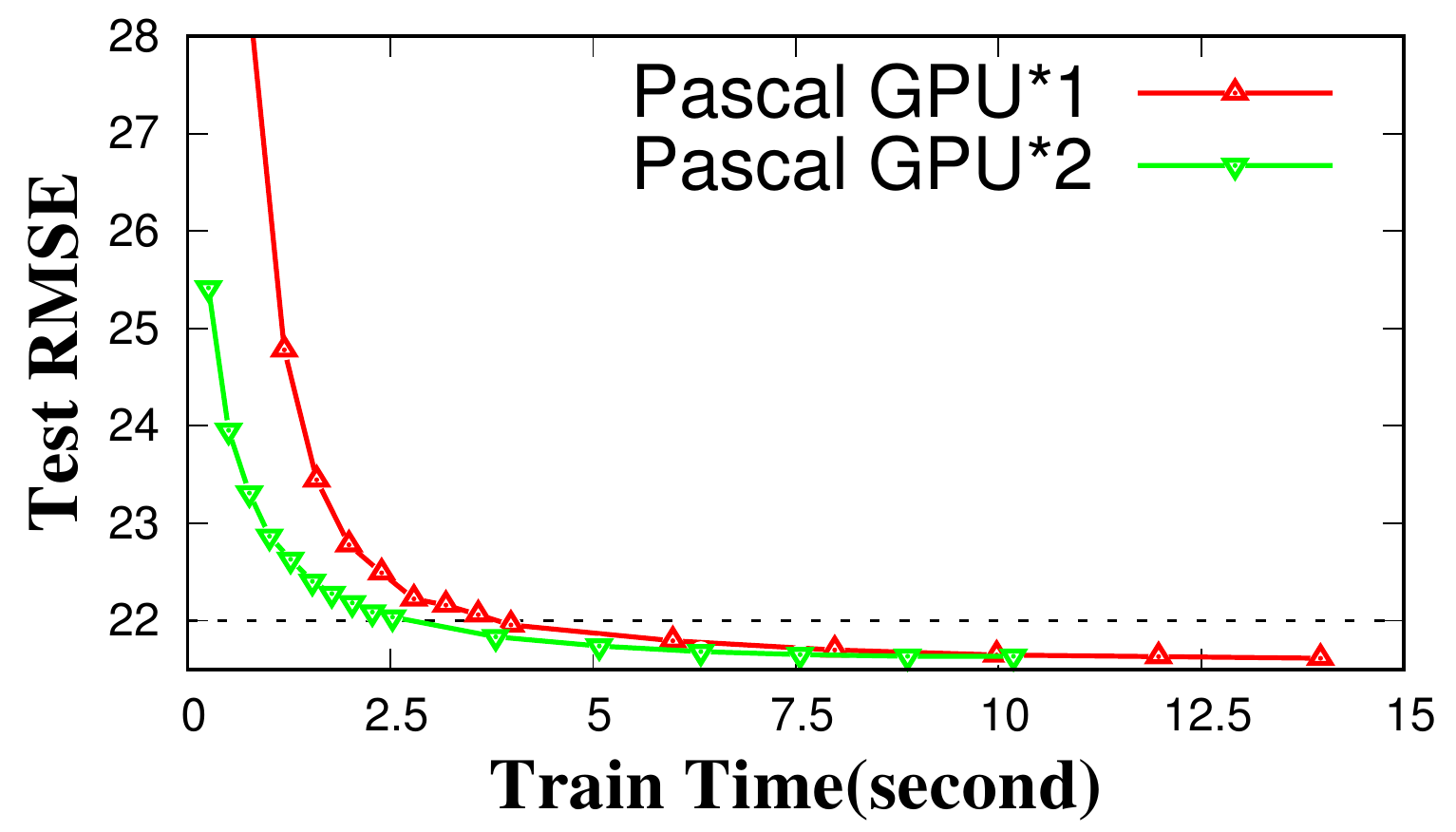}
 \vspace{-0.15cm}
\caption{Convergence of cuMF\_SGD on \textit{Yahoo!Music}: two Pascal GPUs is 1.5X as fast as one.}\label{figure:scale}
 \vspace{-0.3cm}
\end{figure}

System wise, cuMF\_SGD is designed to scale to multiple GPUs. However, algorithmic wise, the scaling is restricted by factors such as problem dimension and number of parallel workers, as discussed earlier in Section~\ref{sec:convergence}. Among the three data sets used in this paper, \textit{Netflix} and \textit{Hugewiki} have very small $n$($20k$, $40k$, receptively), preventing cuMF\_SGD from solving them on multiple GPUs. In comparison, \textit{Yahoo!Music} can be solved on multiple GPUs as the dimension of it $R$ is $1M \times 625k$. We divide its $R$ into $8 \times 8$ blocks and run it with two Pascal GPUs. Figure~\ref{figure:scale} shows the convergence speed. With 2 Pascal GPUs, cuMF\_SGD takes 2.5s to converge to RMSE 22, which is 1.5X as fast as 1 Pascal GPU (3.8s). The reason behind this sub-linear scalability is that, the multi-GPU cuMF\_SGD needs to spend time on CPU-GPU memory transfer so as to synchronize two GPUs.
 \vspace{-0.2cm}

\vspace{-0.3cm}
\section{Related Work}\label{sec:related}

\textbf{Algorithms}. SGD has been widely used to solve matrix factorization~\cite{computer09mf}. Serial SGD can be parallelized to achieve better performance. ALS is naturally easy to parallelize and it can also been used in dense matrix factorization. Coordinate descent is another algorithm to solve matrix factorization~\cite{yu2012scalable,hsieh2011fast}. It updates the feature matrix along one coordinate direction in each step. Our earlier work~\cite{hpdc16wei} focuses on ALS algorithm.

\textbf{Parallel SGD solutions} have been discussed in multi-core~\cite{tist15libmf,libmf++,oh2015fast}, multi-node~\cite{vldb14nomad, teflioudi2012distributed}, MapReduce~\cite{sigkdd11gemulla,li2013sparkler} and parameter-servers~\cite{schelter2014factorbird,cui2014exploiting} settings. Existing works are mostly inspired by Hogwild!~\cite{recht2011hogwild} that allows lock-free update, or matrix-blocking that partitions to avoid conflicts, or a combination of them. LIBMF~\cite{tist15libmf,libmf++} is a representative shared-memory multi-core system. Evaluations have shown that it outperforms all previous approaches with one machine. Although it has been optimized for cache efficiency, it is still not efficient at processing large scale data sets. Moreover, the high complexity of its scheduling policy makes it infeasible to scale to many cores. NOMAD~\cite{vldb14nomad} partitions the data on HPC clusters to improve the cache performance. At the meantime, they propose to minimize the communication overhead. Compared with LIBMF, it has similar performance on one machine and is able to scale to 64 nodes.


Parallelization is also used in coordinate descent ~\cite{yu2012scalable}. Compared with SGD, coordinate descent has lower overhead and runs faster at the first few epochs of training. However, due to the algorithmic limitation, coordinate descent is prone to reach local optima~\cite{tist15libmf} in the later epochs of training.

Compared with CGD and SGD, ALS is inherently easy to parallel, ALS based parallel solutions are widely discussed~\cite{meng2016mllib,zhou2008large,facebook,low2012distributed,bigdataALS}. Our earlier work, cuMF\_ALS~\cite{hpdc16wei} focuses on optimizing ALS to matrix factorization on GPUs. As ALS algorithm is more compute intensive, it runs slower than cuMF\_SGD.

\textbf{GPU solutions}. Prior to our work, ~\cite{cai2012gpu} applies Restricted Boltzmann Machines on GPUs to solve MF. ~\cite{zastrau2012stochastic} implements both SGD and ALS on GPU to solve MF. In contrast, cuMF\_SGD outperforms them because we optimize both memory access and workload scheduling.

\section{Conclusion}\label{sec:conclusion}

Matrix factorization is widely used in recommender systems and other applications. SGD-based MF is limited by memory bandwidth which single and multi-GPU systems cannot efficiently provision. We propose a GPU-based solution, by observing that GPUs offers abundant memory bandwidth and can enjoy fast intra-node connection. We design workload partition and schedule schemes to dispatch tasks insides a GPU and across GPUs, without impacting the randomness required by SGD. We also develop highly-optimized GPU kernels for individual SGD updates. With only one Maxwell or Pascal GPU, cuMF\_SGD runs \textbf{3.1X-28.2X} as fast compared with state-of-art CPU solutions on 1-64 CPU nodes. Evaluations also show that cuMF\_SGD scales well on multiple GPUs in large data sets.

{\scriptsize
\bibliographystyle{ieeetr}
\bibliography{main}
}

\end{document}